\documentclass[letterpaper]{article} 
\usepackage{aaai24}  
\usepackage{times}  
\usepackage{helvet}  
\usepackage{courier}  
\usepackage[hyphens]{url}  
\usepackage{graphicx} 
\usepackage{natbib}  
\usepackage{caption} 
\usepackage{algorithm}
\usepackage{algorithmic}
\usepackage{booktabs}
\usepackage{multirow}
\usepackage{subfigure}
\usepackage{amssymb}
\usepackage{amsmath}

\newcommand{\by}{{\bf y}}
\newcommand{\bz}{{\bf z}}
\newcommand{\bW}{{\bf W}}
\newcommand{\bu}{{\bf u}}
\newcommand{\bb}{{\bf b}}
\newcommand{\bV}{{\bf V}}

\newcommand{\ep}{\epsilon}

\newcommand{\R}{{\mathbb R}}
\newcommand{\Dt}{{\Delta t}}

\newcommand{\bom}{{\bf \omega}}
\usepackage{newfloat}
\usepackage{listings}
\DeclareCaptionStyle{ruled}{labelfont=normalfont,labelsep=colon,strut=off} 
\floatstyle{ruled}
\newfloat{listing}{tb}{lst}{}
\floatname{listing}{Listing}
\title{Neural oscillators for generalization of physics-informed machine learning}
\author{
    Taniya Kapoor\textsuperscript{\rm 1}\equalcontrib,
    Abhishek Chandra\textsuperscript{\rm 2}\equalcontrib,
    Daniel M. Tartakovsky\textsuperscript{\rm 3}, \\
    Hongrui Wang\textsuperscript{\rm 1},
    Alfredo Nunez\textsuperscript{\rm 1},
    Rolf Dollevoet\textsuperscript{\rm 1}
}
\affiliations{
    \textsuperscript{\rm 1}Department of Engineering Structures, Delft University of Technology, The Netherlands\\
    \textsuperscript{\rm 2}Department of Electrical Engineering, Eindhoven University of Technology, The Netherlands\\
    \textsuperscript{\rm 3}Department of Energy Science and Engineering, Stanford University, USA\\

   Correspondence to: t.kapoor@tudelft.nl
}

\usepackage{bibentry}

\begin{document}

\maketitle

\begin{abstract}
A primary challenge of physics-informed machine learning (PIML) is its generalization beyond the training domain, especially when dealing with complex physical problems represented by partial differential equations (PDEs). This paper aims to enhance the generalization capabilities of PIML, facilitating practical, real-world applications where accurate predictions in unexplored regions are crucial. We leverage the inherent causality and temporal sequential characteristics of PDE solutions to fuse PIML models with recurrent neural architectures based on systems of ordinary differential equations, referred to as neural oscillators. Through effectively capturing long-time dependencies and mitigating the exploding and vanishing gradient problem, neural oscillators foster improved generalization in PIML tasks. Extensive experimentation involving time-dependent nonlinear PDEs and biharmonic beam equations demonstrates the efficacy of the proposed approach. Incorporating neural oscillators outperforms existing state-of-the-art methods on benchmark problems across various metrics. Consequently, the proposed method improves the generalization capabilities of PIML, providing accurate solutions for extrapolation and prediction beyond the training data.
\end{abstract}

\section{Introduction}

In machine learning and artificial intelligence, \emph{generalization} refers to the ability of a model to perform on previously unseen data beyond its training domain. This entails prediction of outcomes for a sample \(\mathbf{x}\) that lies outside the convex hull of the training set \(X = \{\mathbf{x_1}, \ldots, \mathbf{x_N}\}\), where $\mathbf{N}$ is the number of training samples \cite{balestriero2021learning}. Current deep-learning models exhibit robust generalization on tasks like image \cite{su2023rethinking}, and speech recognition \cite{chen2023leveraging}, among others \cite{zhou2022domain}. In physical sciences, state-of-the-art deep-learning models, also known as data-driven approaches, learn patterns and correlations from training data but lack intrinsic comprehension of the underlying governing laws of the problem \cite{liu2019predicting, alber2019integrating}. Despite their effective approximation of complex functions and relationships, these data-driven methods face challenges in generalizing to scenarios significantly different from the training distribution, resulting in a physical-agnostic methodology \cite{gu2022neurolight}.

Limitations of data-driven methods, characterized by their inability to adhere to physical laws and their agnosticism towards underlying physics, underscore the need for deep learning models capable of effectively capturing fundamental physical phenomena, such as their structure and symmetry \cite{lee2021machine}. Adopting such learning approaches promises to enhance the generalization capabilities of the model significantly. Consequently, a growing interest has been in embedding physics principles into machine learning to develop physics-aware models such as physics-informed neural networks (PINNs) \cite{raissi2019physics}. PINNs consider mathematical models of the underlying physical process, represented as partial differential equations (PDEs), and integrate them into the loss function during training. 

Despite their popularity, experimental evidence suggests that PINNs might fail to generalize. Minimizing the PDE residual in PINN does not straightforwardly control the generalization error \cite{mishra2023estimates, mishra2022estimates}. Although PINNs and their subsequent enhancement aim to incorporate soft or hard physical constraints for robustness, they often struggle to achieve strong generalization \cite{kim2021dpm, daw2022rethinking, fesser2023understanding}. Hence, simply embedding physical equations into the loss function need not necessarily guarantee genuine physics awareness or robustness beyond the training domain. Ideally, a physics-informed model must reproduce known physics in the training domain and exhibit predictive capabilities for new scenarios while respecting conservation laws and effectively handling variations and uncertainties in real-world applications. Attaining this level of physics awareness remains a crucial challenge in developing dependable and powerful physics-informed machine learning methods \cite{fuks2020limitations, shin2020convergence}.

One way to enhance the extrapolation power of PINNs is to dynamically manipulate the gradients of the loss terms, building upon a gradient-based optimizer  \cite{kim2021dpm}. This method shares similarities with gradient-based techniques employed in domain generalization tasks \cite{wang2022generalizing}. However, one drawback of such methods is the need for training until a specific user-defined tolerance in the loss is achieved, resulting in convergence issues and increased computational costs. We adopt a different strategy to tackle the generalization challenge by leveraging the inherent \emph{causality} present in PDE solutions \cite{wang2022respecting}. Leveraging causality enables us to enhance generalizability by learning the underlying dynamics that preserve the structure and symmetry of the underlying problem.

A recurrent neural network (RNN) might be capable of learning the dynamics owing to its remarkable success in various sequential tasks. Gated architectures, like long short-term memory (LSTM) \cite{hochreiter1997long} and gated recurrent unit (GRU) \cite{cho2014learning}, have been mooted to address the exploding and vanishing gradient problem (EVGP) in vanilla RNNs \cite{pascanu2013difficulty}. However, EVGP can remain a concern as presented by \citeauthor{li2018independently} RNNs with orthogonality constraints on recurrent weight matrices are used to tackle EVGP \cite{henaff2016recurrent, arjovsky2016unitary, wisdom2016full, kerg2019non}. While this strategy alleviates EVGP, it may reduce expressivity and hinder performance in practical tasks \cite{kerg2019non}. We posit that \emph{neural oscillators} \cite{lanthaler2023neural} offers a practical means to achieve high expressibility and mitigate EVGP. Neural oscillators use ordinary differential equations (ODEs) to update the hidden states of the recurrent unit, enabling efficient dynamic learning. 

This paper introduces a new approach to address the generalization challenge. It employs a physics-informed neural architecture that learns the underlying dynamics in the training domain, followed by a neural oscillator to exploit the causality and learn temporal dependencies between the solutions at subsequent time levels. This extension of a physics-informed architecture helps increase the accuracy of a generalization task since neural oscillators carry a hidden state that retains information from previous time steps, enabling the model to capture and leverage temporal dependencies in the data.

We consider two different neural oscillators: \emph{coupled oscillatory recurrent neural network} (CoRNN) \cite{rusch2020coupled} and \emph{long expressive memory} (LEM) \cite{rusch2021long}. Both methods use a coupled system of ODEs to update the hidden states. We ascertain the relative performance of these two oscillators on three benchmark nonlinear problems: viscous Burgers equation, Allen--Cahn equation, and Schr\"{o}dinger equation. Additionally, we evaluate the performance of our method in generalizing a solution for the Euler--Bernoulli beam equation. To showcase the performance of the proposed framework for higher-dimensional PDEs, we performed an experiment on 2D Kovasznay flow as presented in supplementary material \textbf{SM}\S C provided at \cite{kapoor2023neuralarxiv}

The remainder of the manuscript is structured as follows. The ``Related Work" section provides an overview of pertinent literature and recent studies related to the current work. In the ``Method" section, our approach for enhancing the generalization of physics-informed machine learning through integration with a neural oscillator is explained in detail. Our method is validated through a series of numerical experiments in the ``Numerical Experiments" section. Finally, key findings and implications of this study are collated in the ``Conclusions" section.

\section{Related Work}


\begin{figure*}[t]
\begin{center}
\includegraphics[width=0.95\textwidth]{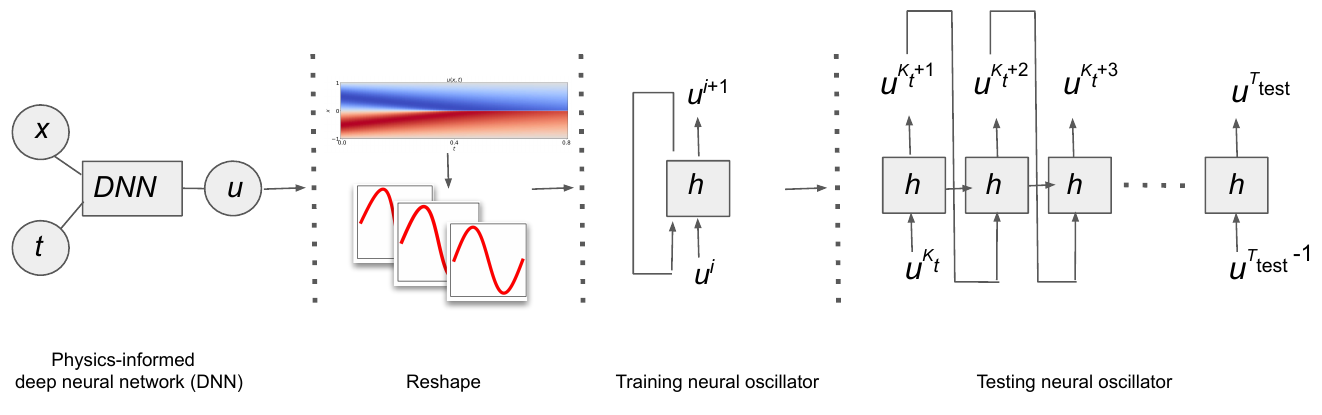}
\caption{The proposed framework in which a physics-informed architecture (e.g., PINN or its variants) learns a solution in the convex hull \(X_1\). After reshaping, these solutions are represented sequentially and processed by one of the neural oscillators. The neural oscillator is finally tested in the convex testing hull \(X_2\), where the output of the last prediction step is the input for the next prediction step. Here, $1 \leq i \leq k_t$, $i \in \mathbb{Z}$, and $h = [y, z]$. The dotted lines separate different stages of training and testing the framework.}
\label{fig:proposed_architecture}
\end{center}
\end{figure*}

\subsubsection{PIML}
Our research aims to advance physics-informed models, a subset of machine learning techniques that address physical problems formulated as PDEs. PIML encompasses a range of methodologies, including physics-informed \cite{karniadakis2021physics}, physics-based \cite{cuomo2022scientific}, physics-guided \cite{daw2022physics}, and theory-guided \cite{cuomo2022scientific} approaches. The review papers \cite{karniadakis2021physics, cuomo2022scientific} provide a comprehensive overview of progress in PIML. Recently, PIML has demonstrated considerable utility in scientific and engineering disciplines, encompassing fluid dynamics \cite{raissi2020hidden} and materials science \cite{zhang2022analyses}, among others. Our primary focus is to improve PIML variants that integrate governing equations into the loss function during training to foster generalization, which involves advancing PINNs and their variations, such as causal PINNs \cite{wang2022respecting}, and self-adaptive PINNs \cite{mcclenny2023self}.

\subsubsection{Domain generalization}
Domain generalization focuses on training models to effectively handle unseen domains with diverse data distributions, even when trained on data from related but distinct domains \cite{zhou2022domain, wang2022generalizing}. In contrast, domain adaptation involves transferring knowledge from a labeled source domain to an unlabeled or partially labeled target domain, assuming access to some labeled data in the target domain \cite{shen2018wasserstein}. Our research shares the core principles with these fields but differs in that we learn \emph{exclusively} from a single training set without using multiple domains, as in domain generalization, or having access to any target domain data, as in domain adaptation. Moreover, we do not employ any transfer learning techniques. Our task is to train \emph{solely} on the training set and \emph{directly deploy} the trained model on the test region.

\subsubsection{Generalization in PIML}
Despite limited research on the generalization of physics-informed models, some studies have specifically focused on generalizing PINNs. One noteworthy approach is the dynamic pulling method (DPM) \cite{kim2021dpm}, which utilizes a gradient-based technique to extend the solution of nonlinear benchmark problems beyond the trained convex hull \(X\), focusing on generalizing solutions in the temporal domain. Other investigations have centered on generalizing the parameter space for parametric PDEs, employing techniques like curriculum learning, sequence-to-sequence learning \cite{krishnapriyan2021characterizing} and incremental learning \cite{dekhovich2023ipinns}. However, these approaches involve training and testing \emph{within the convex hull} of the parameter space, which differs from the focus and approach to our work.

\subsubsection{Neural oscillators}
Oscillator networks are ubiquitous in natural and engineering systems, exemplified by pendulums (classical mechanics) and heartbeats (biology). A growing trend involves building RNN architectures based on ODEs and dynamical systems \cite{chen2018neural, rubanova2019latent, chang2019antisymmetricrnn, rusch2021unicornn}. Recent research has abstracted the fundamental nature of functional brain circuits as networks of oscillators, constructing RNNs using simpler mechanistic systems represented by ODEs while disregarding complex biological neural function details. Driven by the \emph{long-term memory} of these oscillators and inspired by the \emph{universal approximation property} \cite{lanthaler2023neural}, our goal is to integrate them with physics-informed models to enhance generalization.

\section{Method}
The proposed framework comprises a feedforward neural network informed by physics (such as PINN, causal PINN, self-adaptive PINN, or any other physics-guided architecture), followed by a neural oscillator. For example, we combine PINN with the coupled oscillatory recurrent neural network (CoRNN) or the long expressive memory (LEM) model. The output of the PINN serves as input to the oscillator. The PINN learns a solution within a convex training hull \(X_1 = D \times T\), where \(D \in \mathbb{R}^{d}\) is the $d$-dimensional spatial domain and \(T \in \mathbb{R}\) is the temporal domain of the PDE. In our experiments, $d=1$.

The neural oscillator processes the PINN's output as sequential data and predicts solutions within a different convex testing hull \(X_2\). The hulls are distinct, \(X_1\) and \(X_2\),  and \(X_2 \nsubseteq X_1\). For example, \(X_2 = D \times T^{'}\), where \(D\) is the same spatial domain but \(T^{'} \in \mathbb{R}\) is the extrapolated temporal domain with $\inf(T^{'}) \geq \sup(T)$, which implies that testing is performed on time \(t^{'} \in T^{'} \geq t \in T\). 

The PINN maps the input space \(X_1\) onto the solution space $\mathcal{U}$, such that a solution of the PDE $u \in \mathcal{U}$. This mapping enables learning the evolution of $u$ from a given initial condition. The abstract formulation of an operator $\mathcal{N}$ incorporating the PDE and initial and boundary conditions is
\begin{equation}
\label{eq1}
\mathcal{N}(u) = f,
\end{equation}
where $f$ is the source term. The loss function of an abstract PINN is formed by minimising the residuals of \eqref{eq1} along with the available data on boundaries and at the initial time.

Following the PINN training on \(X_1\), its testing is conducted on $k_t$ uniform time steps in $T$ and $k_x$ uniform locations in $D$ making a total of $k_t \cdot k_x$ testing points within \(X_1\). The solution obtained from the PINN is reshaped to be further fed into the neural oscillator (Fig.~\ref{fig:proposed_architecture}). 

Conventional feed-forward neural networks lack explicit mechanisms to learn dependencies among outputs, presenting a fundamental \emph{challenge in handling temporal relationships}. To mitigate this challenge, recurrent neural architectures preserve a hidden state to retain information from previous time steps, thereby improving sequence learning. We employ neural oscillators to treat the PINN's outputs as a sequence. The motivation arises from feed-forward neural networks, where all outputs are independent, whereas sequence learning requires capturing temporal dependencies. Neural oscillators capture these dependencies through feedback loops and hidden states, enabling information propagation and temporal dependency capture. 

While training an oscillator, its hidden states are updated using the current input and the previous hidden states, akin to vanilla RNNs. The fundamental distinction between vanilla or gated RNNs and neural oscillators lies in the hidden state update methodology. In neural oscillators, these updates are based on systems of ODEs, in contrast to algebraic equations used in typical RNNs. When employing CoRNN, the hidden states are updated through the second-order ODE
\begin{equation}
\label{eq:ode1}
\by^{\prime \prime} = \sigma\left(\bW \by +  \boldsymbol{\mathcal{W}} \by^{\prime} + \bV \bu + \bb \right) -\gamma \by - \ep \by^{\prime}.
\end{equation}
Here, $\by = \by (t) \in \R^m$ is the hidden state of the RNN with weight matrices $\bW, \boldsymbol{\mathcal{W}} \in \R^{m \times m}$ and $\bV \in \R^{m \times k_x}$; $t$ corresponds to the time levels at which the PINN's testing has been performed; $\bu = \bu(t) \in  \R^{k_x}$ is the PINN solution; 
$\bb \in \R^m$ is the bias vector; and $\gamma,\ep > 0$ are the oscillatory parameters. We set the activation function $\sigma: \R \mapsto \R$ to $\sigma (u) = \tanh(u)$. Introducing $\bz = \by^{\prime}(t) \in \R^m$, we rewrite~\eqref{eq:ode1} as the first-order system
\begin{equation}
\label{eq:ode}
\by^{\prime} = \bz, \quad
\bz^{\prime}= \sigma\left(\bW \by +  \boldsymbol{\mathcal{W}} \bz + \bV \bu + \bb \right)  - \gamma \by - \ep \bz.
\end{equation}

We use an explicit scheme with a time step $0 < \Dt < 1$ to  discretize these ODEs, 
\begin{equation}
\label{eq:brnn}
\begin{aligned}
\by_n &= \by_{n-1} + \Dt \bz_n,\\
\bz_n &= \bz_{n-1} + \Dt \sigma\left(\bW\by_{n-1} +  \boldsymbol{\mathcal{W}} \bz_{n-1} + \bV \bu_{n} + \bb \right) \\ &\quad -\Dt \gamma \by_{n-1} -  \Dt \ep \bz_{\bar{n}}.
\end{aligned}
\end{equation}

Similarly, LEM updates the hidden states by solving the ODEs 
\begin{equation}
    \label{eq:lemode}
    \begin{aligned}
    \by^{\prime} &= \hat{\sigma}(\bW_2\by + \bV_2\bu + \bb_2)\odot [\sigma(\bW_y \bz + \bV_y \bu + \bb_y) - \by] \\
    \bz^{\prime} & = \hat{\sigma}(\bW_1\by + \bV_1\bu + \bb_1)\odot [\sigma(\bW_z \by + \bV_z \bu + \bb_z) - \bz] 
    \end{aligned}
\end{equation}
In addition to previously defined quantities, $\bW_{1,2}, \bW_{y,z} \in \R^{m\times m}$ and $\bV_{1,2}, \bV_{y,z} \in \R^{m\times k_x}$ are the weight matrices; $\bb_{1,2}$ and $\bb_{y,z} \in \R^{m}$ are the bias vectors; $\hat{\sigma}$ is the sigmoid activation function; and $\odot$ refers to the componentwise product of vectors. A discretization of~\eqref{eq:lemode} similar to CoRNN yields
\begin{equation}
    \label{eq:lem}
    \begin{aligned}
    {\bf \Dt}_n &= \Dt\hat{\sigma}(\bW_1\by_{n-1} + \bV_1\bu_{n} + \bb_1)  \\ 
    {\bf \overline{\Dt}}_n &= \Dt\hat{\sigma}(\bW_2\by_{n-1} + \bV_2\bu_{n} + \bb_2)\\
    \bz_n &= (1 - {\bf \Dt}_n) \odot  \bz_{n-1} \\ &\quad + {\bf \Dt}_n \odot \sigma(\bW_z \by_{n-1} + \bV_z \bu_{n} + \bb_z) \\
    \by_n &= (1 - {\bf \overline{\Dt}}_n)  \odot \by_{n-1} \\ &\quad + {\bf \overline{\Dt}}_n\odot\sigma(\bW_y \bz_{n} + \bV_y \bu_{n} + \bb_y).
\end{aligned}
\end{equation}
Both CoRNN and LEM are augmented with a linear output state $\bom_n \in \R^{k_x}$ with $\bom_n =\mathcal{Q} \by_n$ and $\mathcal{Q} \in \R^{k_x\times m}$.

We train the PINN and the neural oscillator separately to leverage the \emph{resolution-invariance} property of physics-informed learning during training. While neural oscillators require evenly spaced data, a PINN can be trained discretization-invariantly, allowing flexibility in handling multi-resolution data, such as using different sampling techniques \cite{daw2022rethinking}. The PINN is trained until a predefined epoch or until its validation error stabilizes in consecutive epochs and is then employed in inference to generate training data for the oscillator. Subsequently, the oscillator learns a mapping between the PINN outputs from one-time level to the next, forming a sequential relationship.

\begin{table*}
\setlength{\tabcolsep}{2pt}
\centering
\scriptsize 
\caption{The generalization accuracy in terms of the relative errors in the L2-norm, the explained variance error, the max error, and the mean absolute error for nonlinear benchmark PDEs. Higher (or lower) values are preferred, corresponding to $\uparrow$ (or $\downarrow$).}\label{tbl:1}
\begin{tabular}{|c|c|c|c|c|c|c|c|c|c|c|c|c|}
\hline
\multirow{2}{*}{PDE} & \multicolumn{3}{c|}{L2-norm $(\downarrow)$} & \multicolumn{3}{c|}{Explained variance score $(\uparrow)$} & \multicolumn{3}{c|}{Max error $(\downarrow)$} & \multicolumn{3}{c|}{Mean absolute error $(\downarrow)$} \\ \cline{2-13}
          & DPM & CoRNN & LEM & DPM & CoRNN & LEM & DPM & CoRNN & LEM & DPM & CoRNN & LEM \\ \hline
Vis. Burgers & 0.083  & 0.0044 &\textbf{0.0001}&  0.621 & 0.9955 &\textbf{0.9998}  &1.534  &0.1035  &\textbf{0.0246}& 0.277  & 0.0222 &\textbf{0.0035}  \\ \hline
Allen--Cahn & 0.182  &0.0051  & \textbf{0.0049} &  0.967  &0.9954  & \textbf{0.9956}  & 0.836  & 0.3201 & \textbf{0.1376} & 0.094  &0.0356  & \textbf{0.0348}  \\ \hline
Schr\"{o}dinger & 0.141   & 0.0426 & \textbf{0.0034}  & -3.257  & 0.9250 & \textbf{0.9944}  & 3.829 & 0.6596 & \textbf{0.0948} & 0.868 & 0.9250 & \textbf{0.0281}  \\ \hline
\end{tabular}
\end{table*}

\begin{figure}[t]
    \centering
    \subfigure[Reference Solution]{\label{fig:visc_pinn_ref}\includegraphics[width=0.23\textwidth]{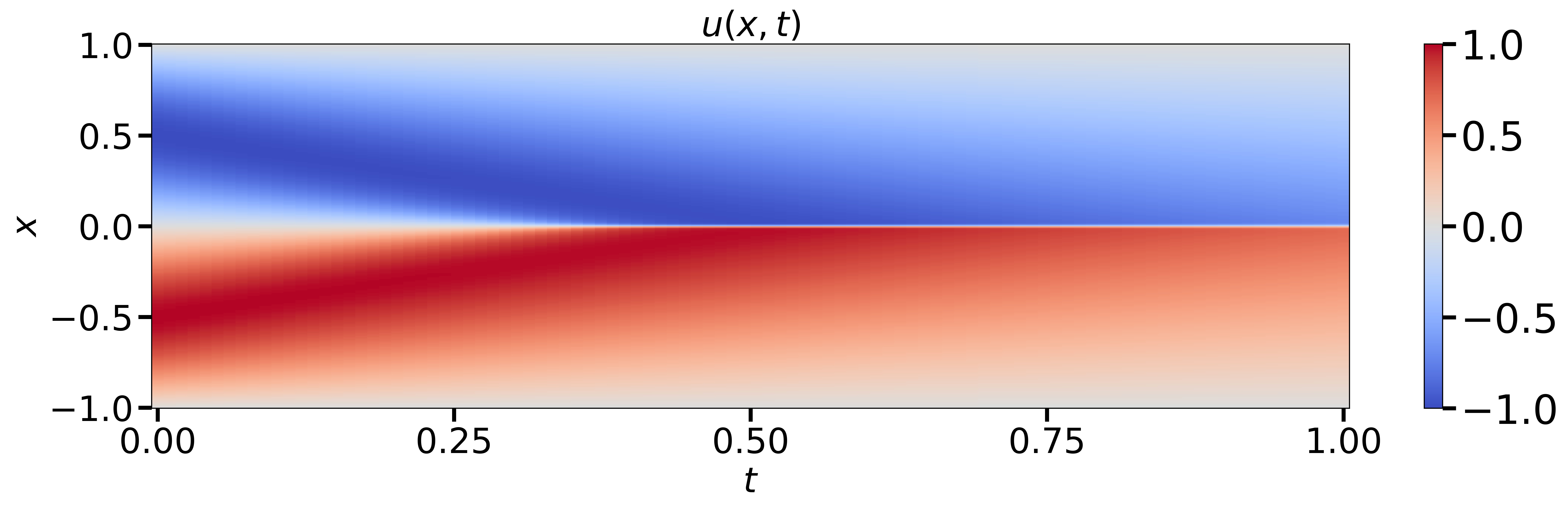}}\hfill
    \subfigure[GRU]{\label{fig:Burger_GRU_contour}\includegraphics[width=0.23\textwidth]{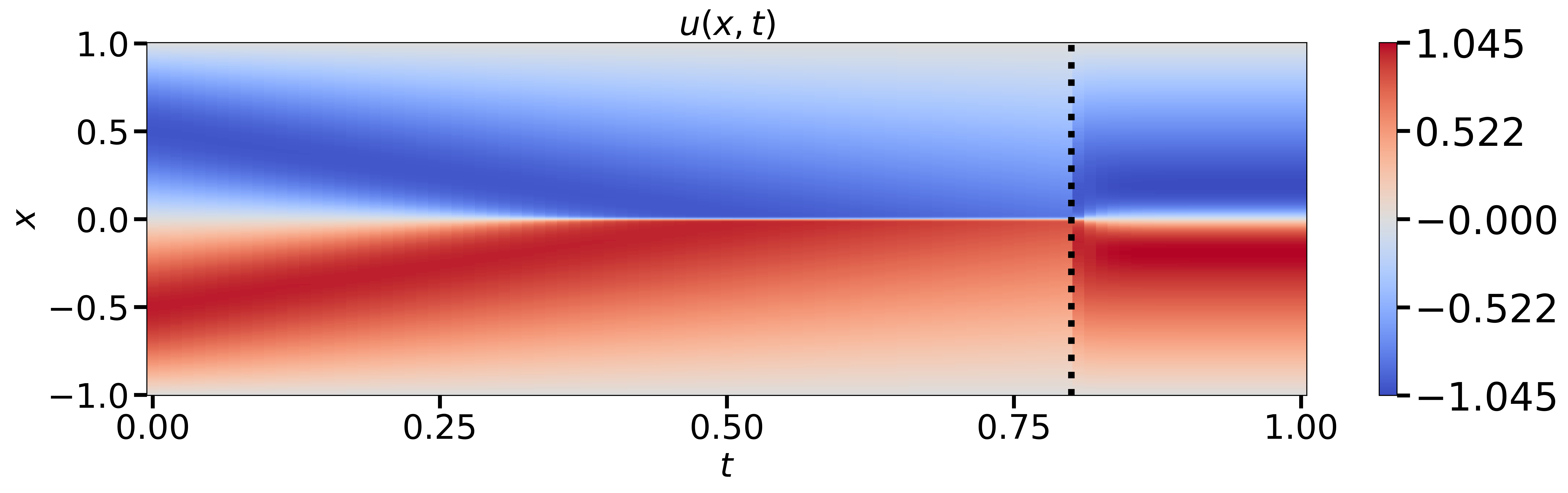}}\hfill
    \subfigure[CoRNN]{\label{fig:Burger_CoRNN_contour}\includegraphics[width=0.23\textwidth]{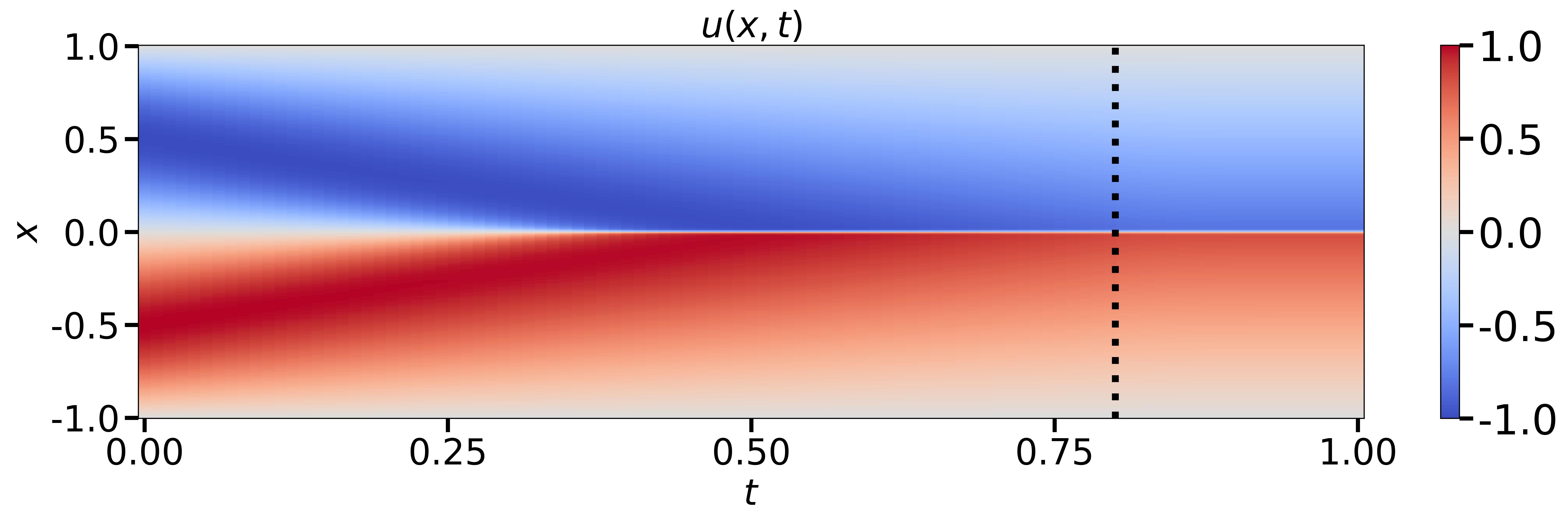}}\hfill
    \subfigure[LEM]{\label{fig:Burger_LEM_contour}\includegraphics[width=0.23\textwidth]{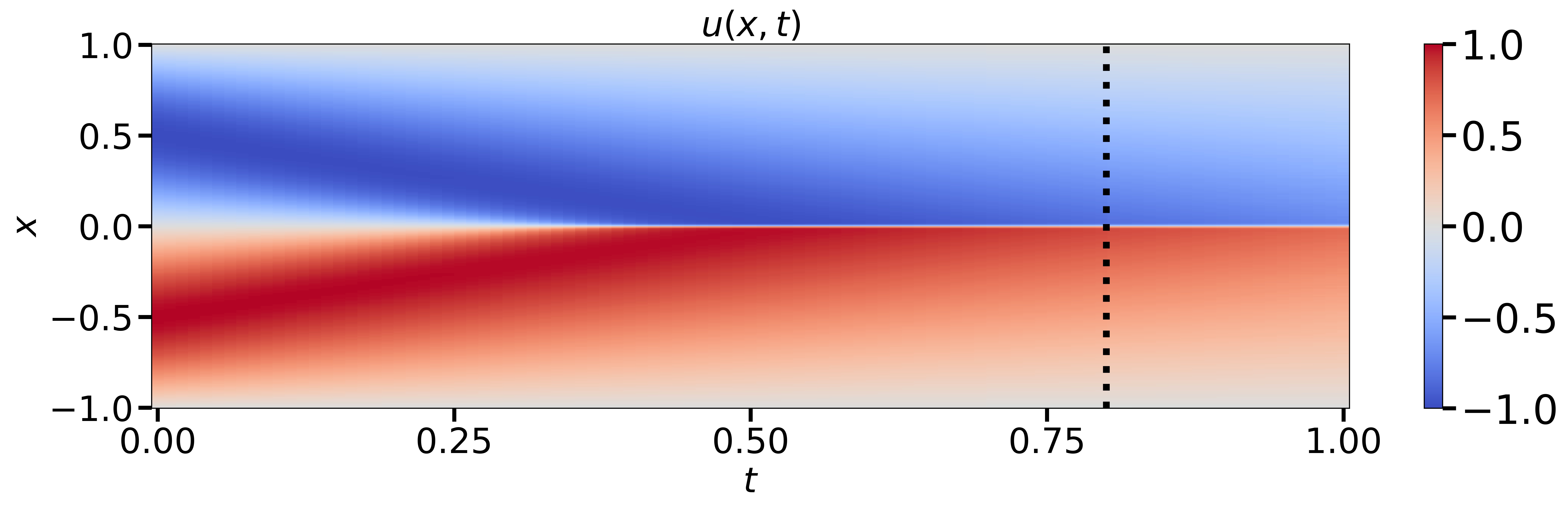}}\\
    \subfigure[GRU]{\label{fig2e}\includegraphics[height=2.5cm, width=2.5cm]{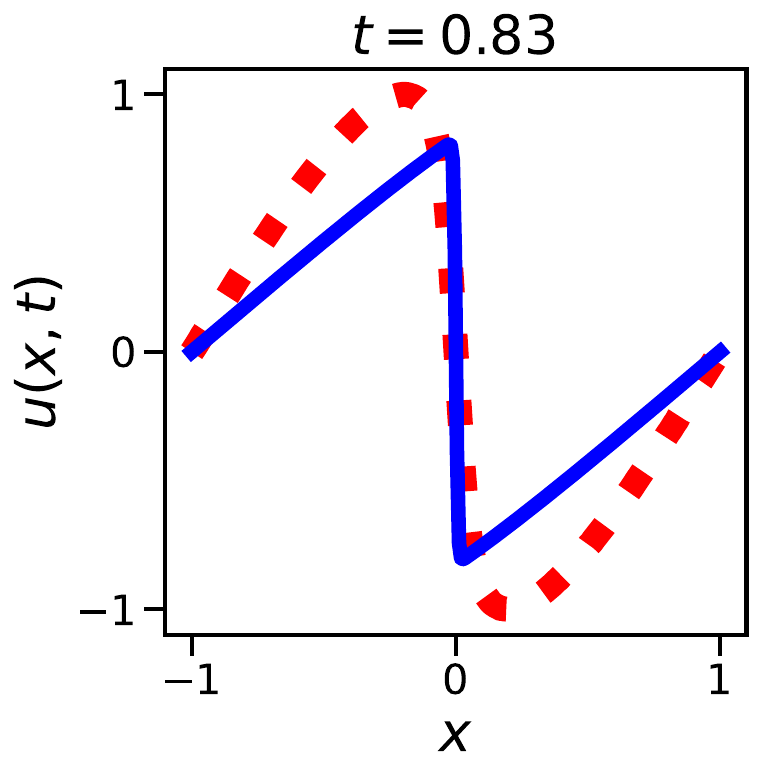}}\hfill
      \subfigure[CoRNN]{\label{fig2f}\includegraphics[height=2.5cm, width=2.5cm]{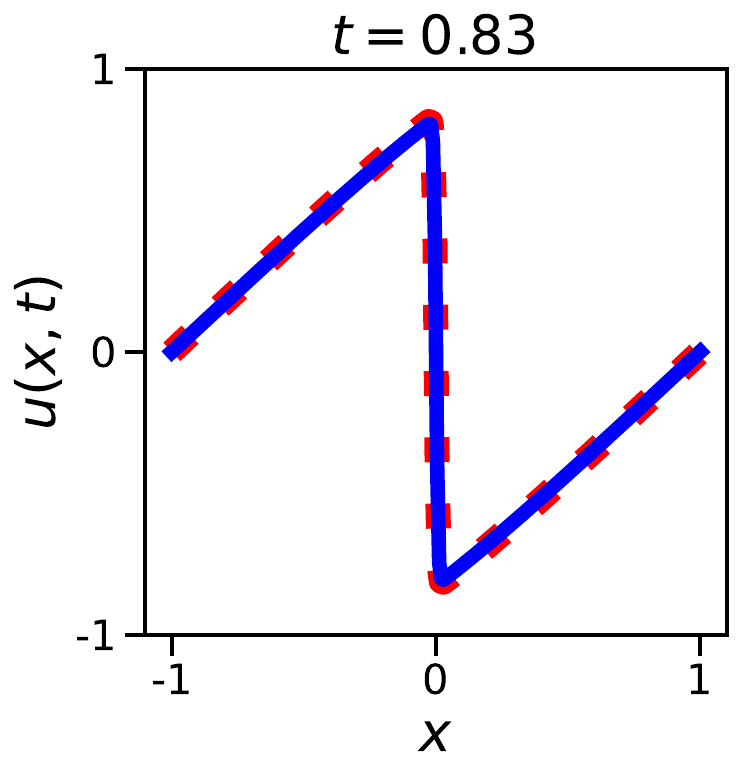}} \hfill
      \subfigure[LEM]{\label{fig2g}\includegraphics[height=2.5cm, width=2.5cm]{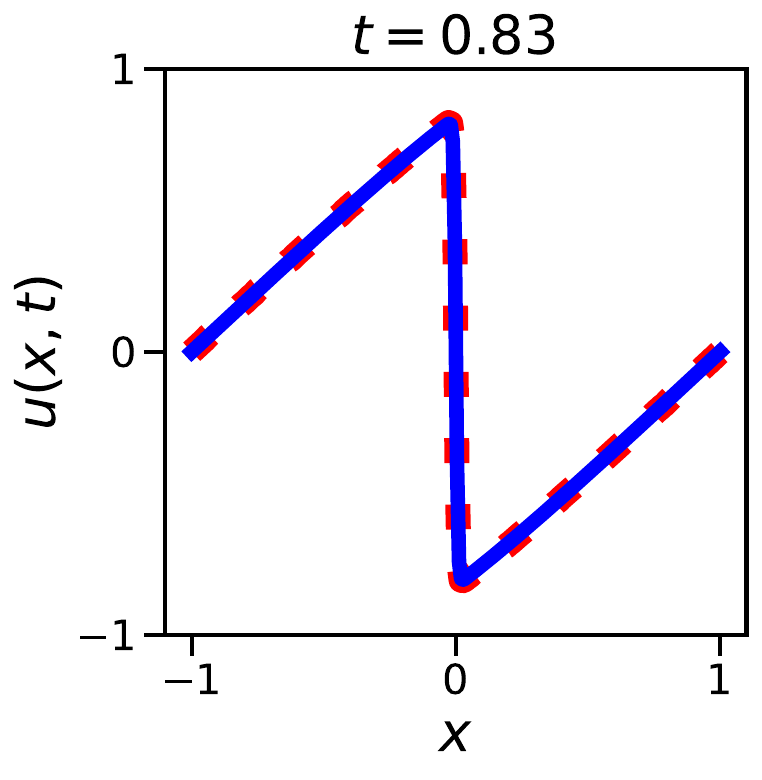}} \hfill
    \subfigure[GRU]{\includegraphics[height=2.5cm, width=2.5cm]{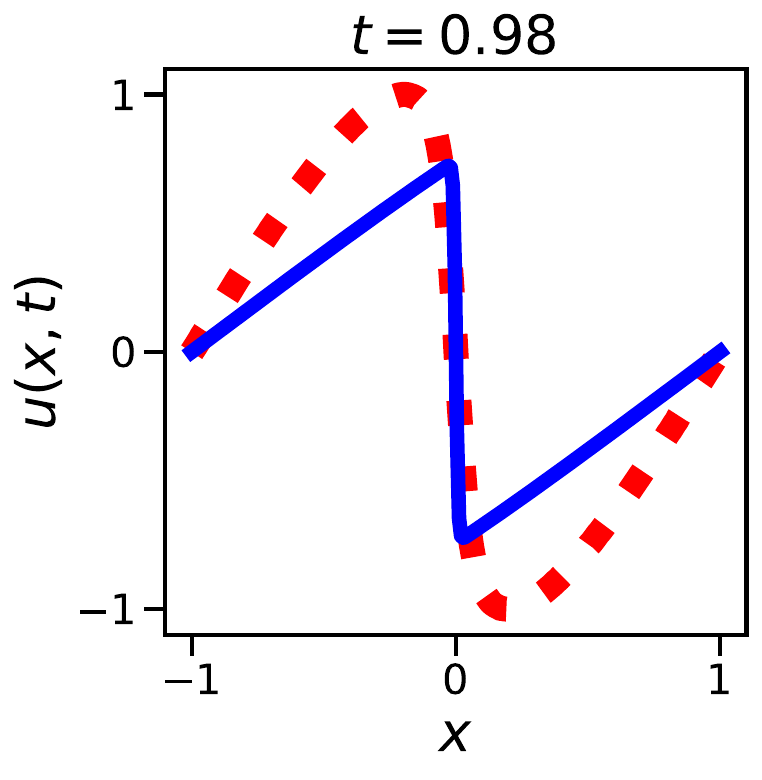}}\hfill
     \subfigure[CoRNN]{\includegraphics[height=2.5cm, width=2.5cm]{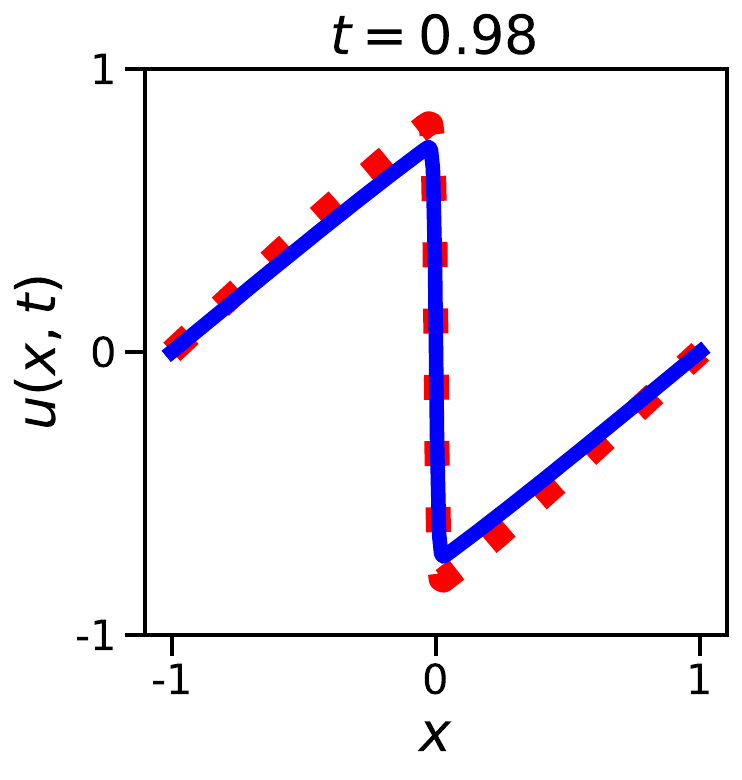}}\hfill
        \subfigure[LEM]{\label{fig2j}\includegraphics[height=2.5cm, width=2.5cm]{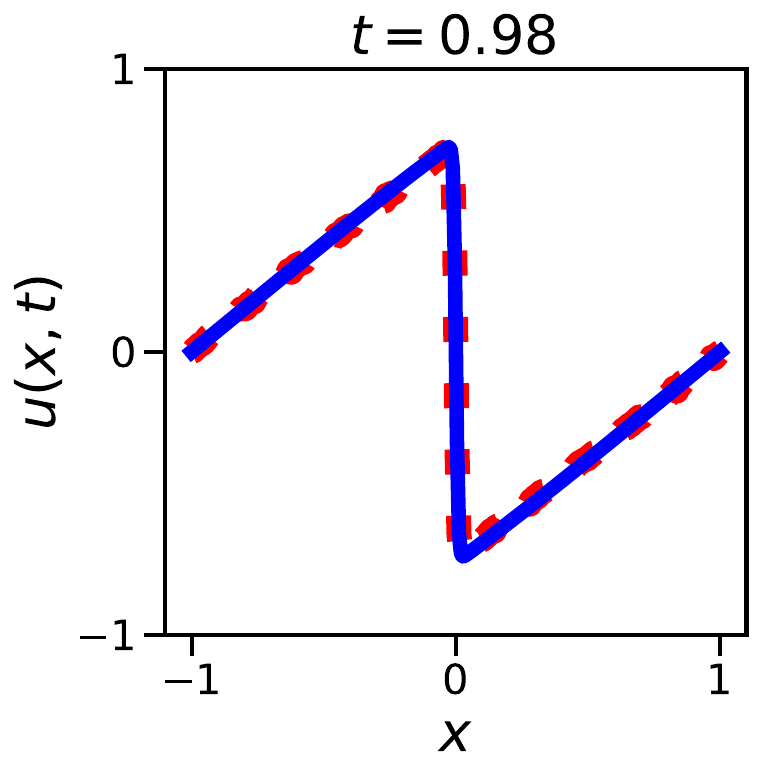}}\hfill
    \caption{Top two rows: the complete reference solution and predictions for viscous Burgers equation. The black vertical line delineates the region before which the PINN has been trained. The region after the black vertical line represents the generalization domain. The meaning of the vertical line remains the same in the following figures. Bottom: the solution snapshots at $t=\{0.83,0.98\}$ obtained in the generalization region, where blue represents the reference solution, and red refers to the recurrent method. The colors are used consistently for the following figures.}
    \label{fig:burgers}
\end{figure}

\begin{figure}[t]
    \centering
    \subfigure[Reference Solution]{\label{fig:AC_pinn_ref}\includegraphics[width=0.23\textwidth]{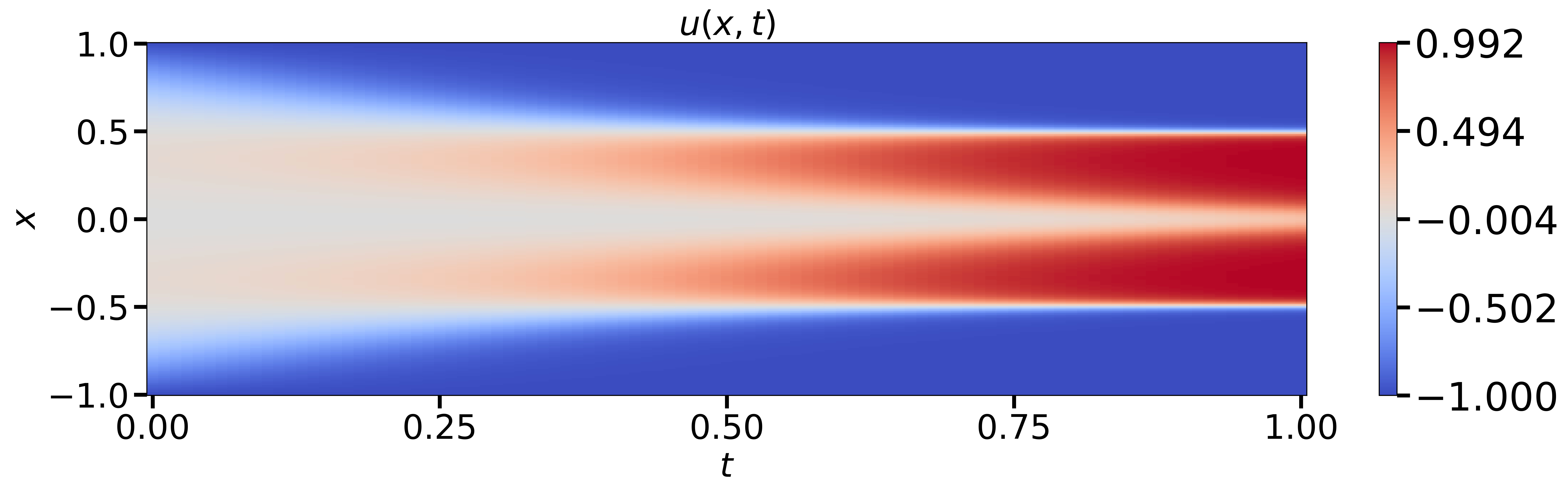}}\hfill
    \subfigure[GRU]{\label{fig:AC_GRU_contour}\includegraphics[width=0.23\textwidth]{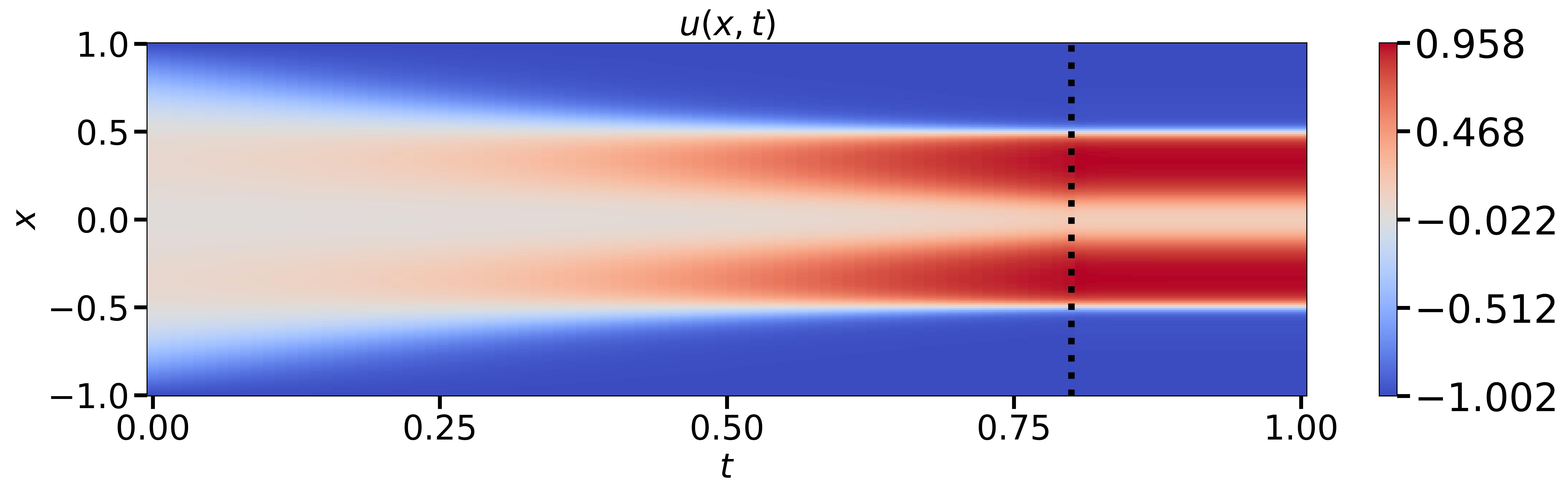}}\hfill
    \subfigure[CoRNN]{\label{fig:AC_CoRNN_contour}\includegraphics[width=0.23\textwidth]{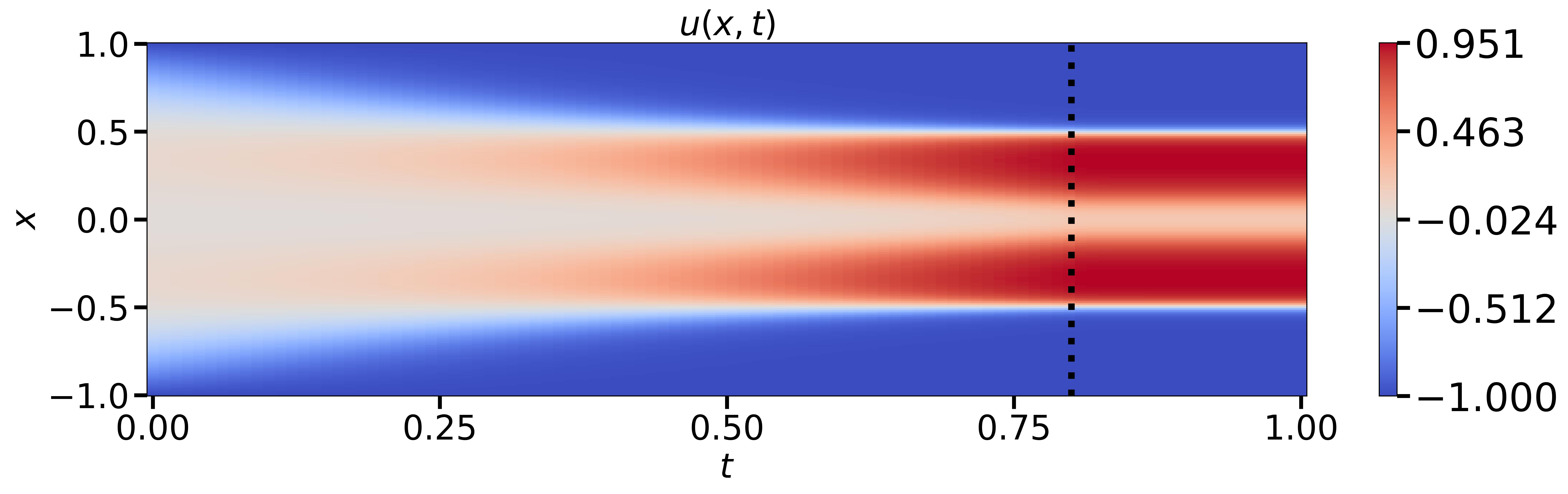}}\hfill
    \subfigure[LEM]{\label{fig:AC_LEM_contour}\includegraphics[width=0.23\textwidth]{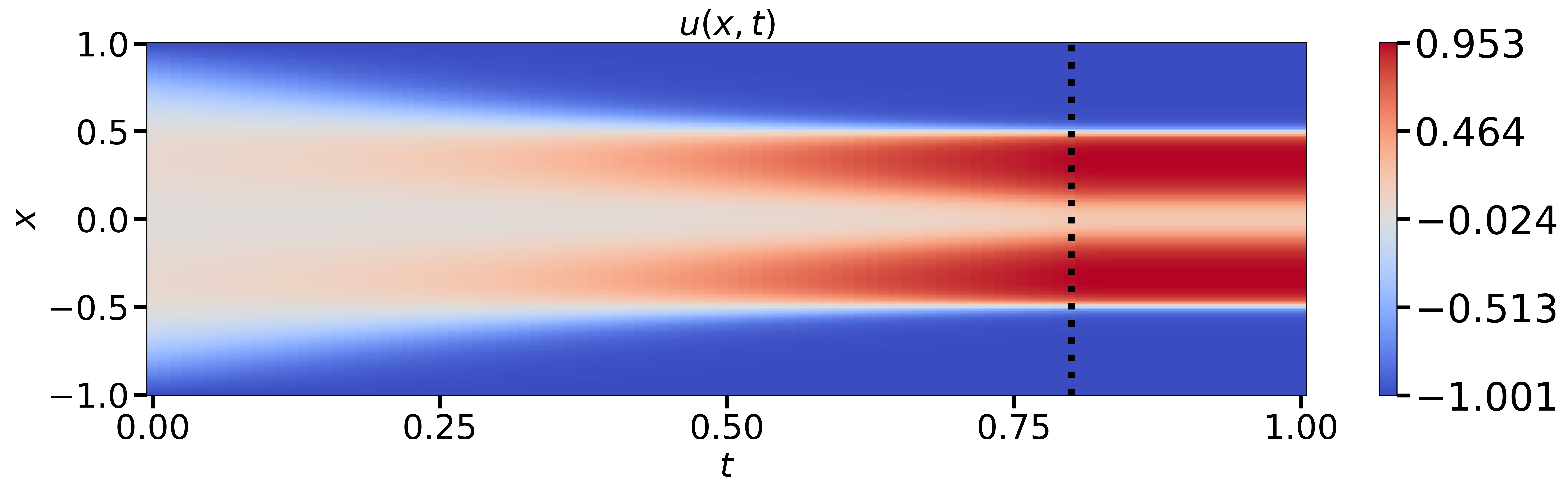}}\\
    \subfigure[GRU]{\label{fig3e}\includegraphics[height=2.5cm, width=2.5cm]{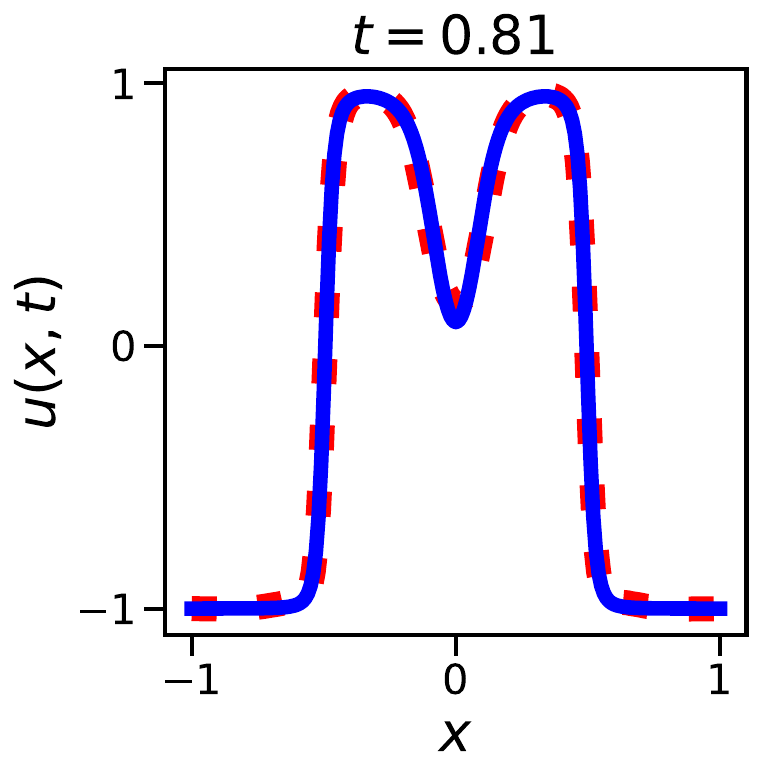}}\hfill
      \subfigure[CoRNN]{\includegraphics[height=2.5cm, width=2.5cm]{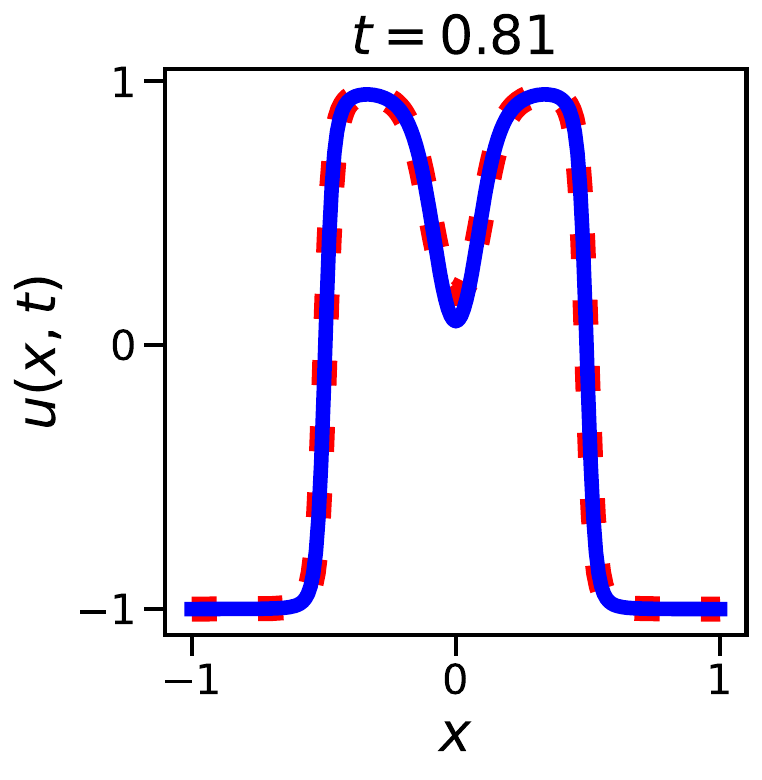}} \hfill
      \subfigure[LEM]{\label{fig3g}\includegraphics[height=2.5cm, width=2.5cm]{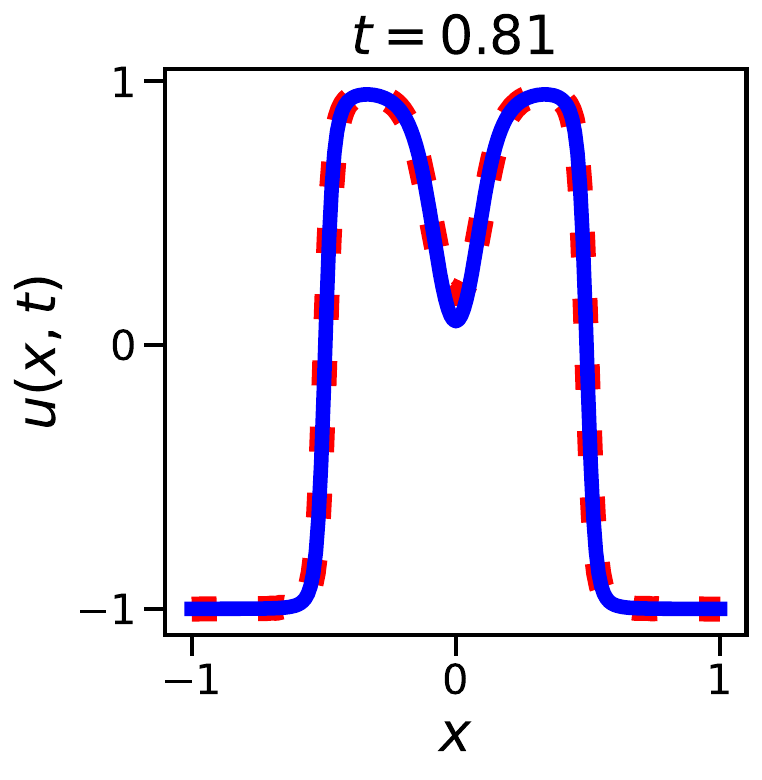}} \hfill
    \subfigure[GRU]{\label{fig3h}\includegraphics[height=2.5cm, width=2.5cm]{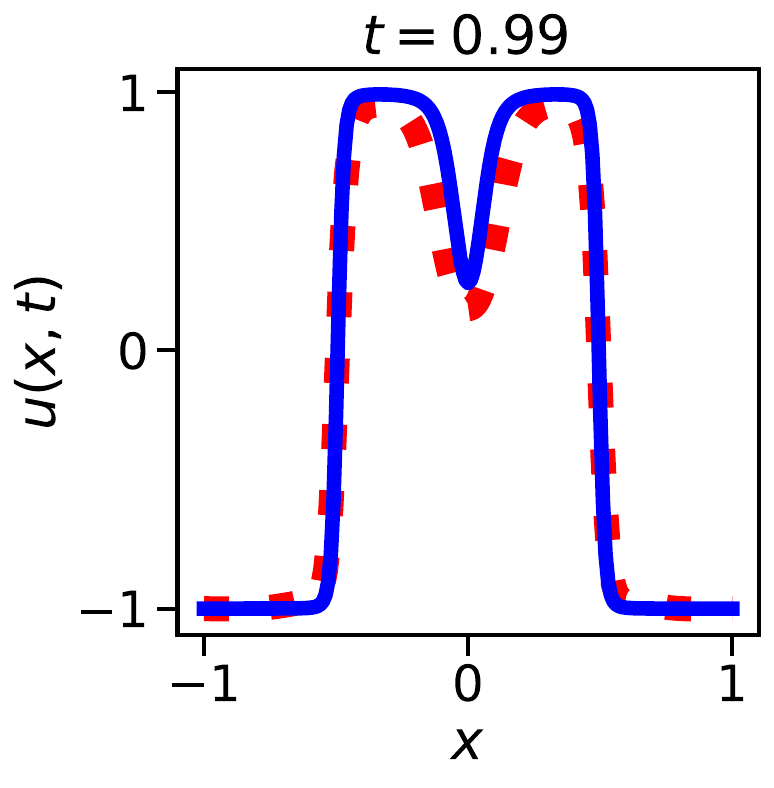}}\hfill
     \subfigure[CoRNN]{\includegraphics[height=2.5cm, width=2.5cm]{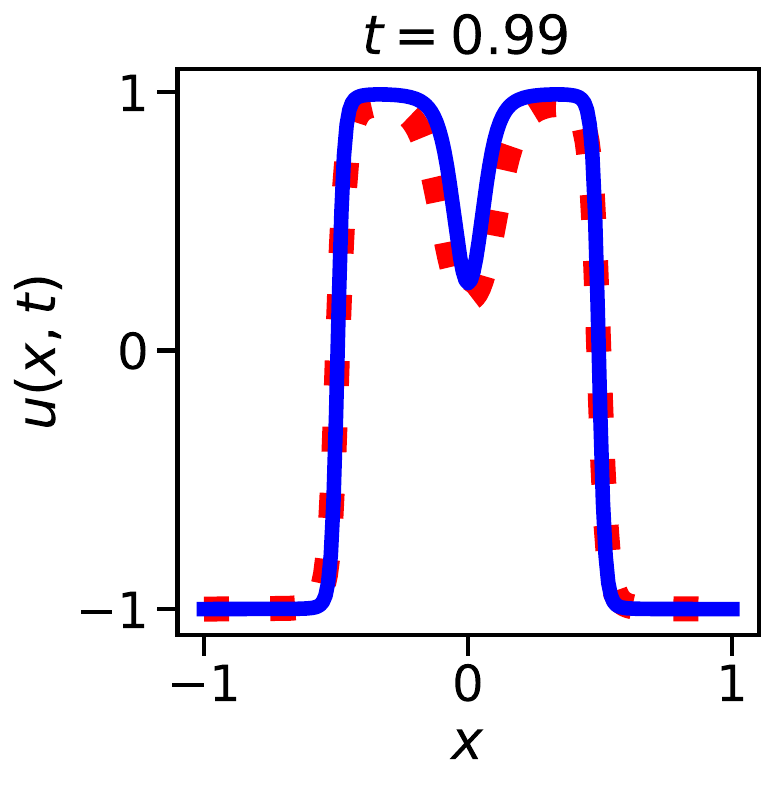}}\hfill
        \subfigure[LEM]{\label{fig3j}\includegraphics[height=2.5cm, width=2.5cm]{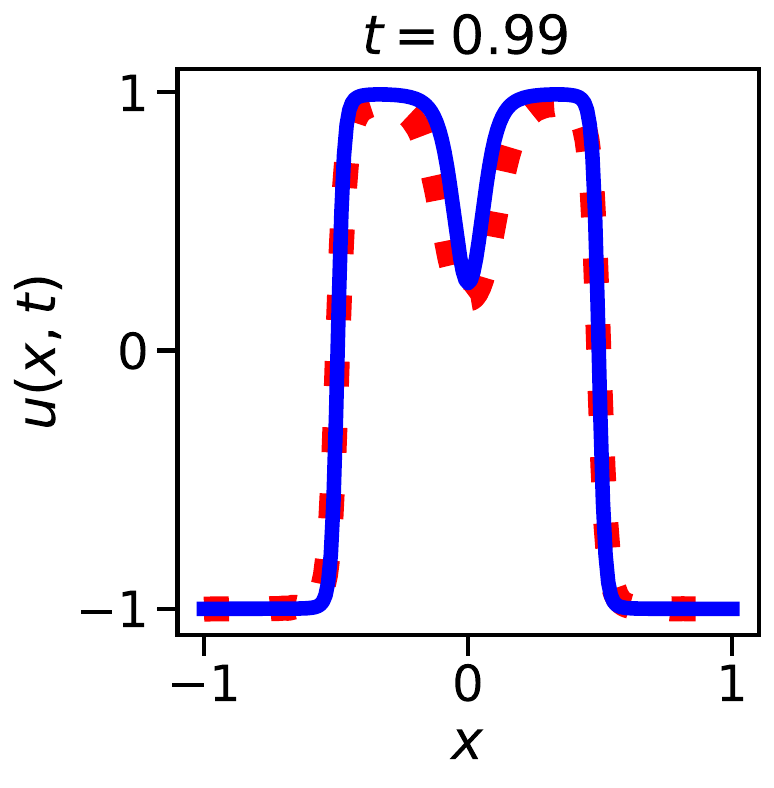}}\hfill
    \caption{Top two rows: the complete reference solution and predictions for the Allen-Cahn equation. Bottom: the solution snapshots at $t=\{0.81, 0.99\}$ obtained in the generalization region.}
    \label{fig:AC}
\end{figure}

\begin{figure}[t]
    \centering
    \subfigure[Reference Solution]{\label{fig:SC_pinn_ref}\includegraphics[width=0.23\textwidth]{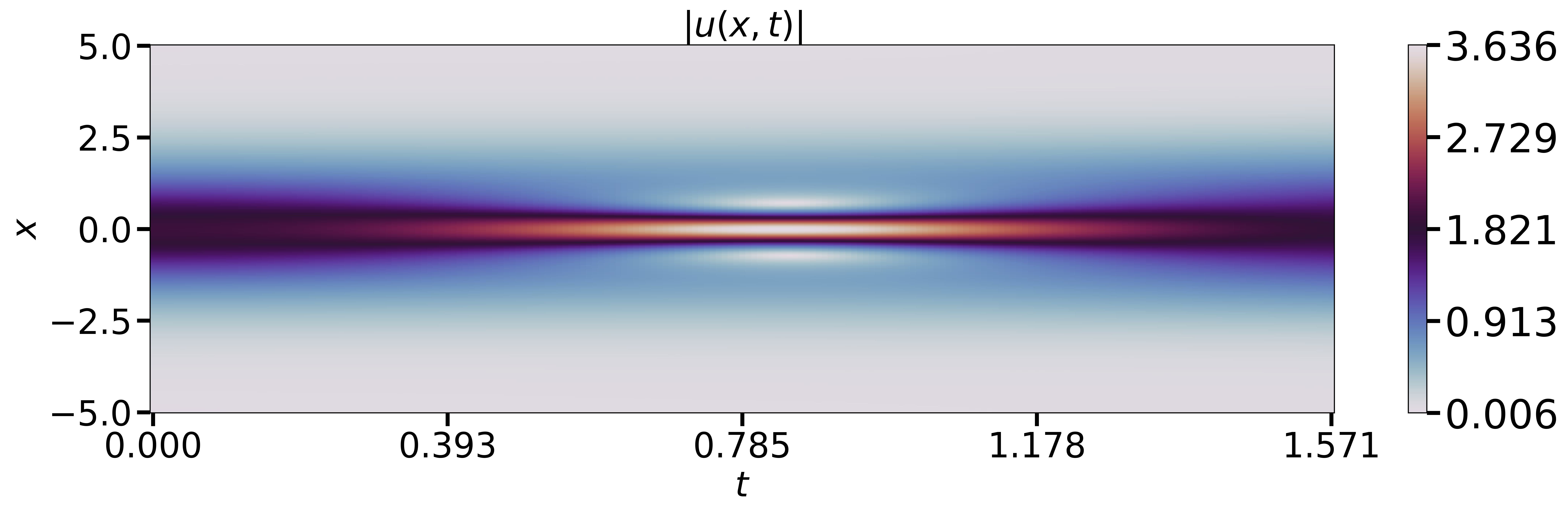}}\hfill
    \subfigure[GRU]{\label{fig:SC_GRU_contour}\includegraphics[width=0.23\textwidth]{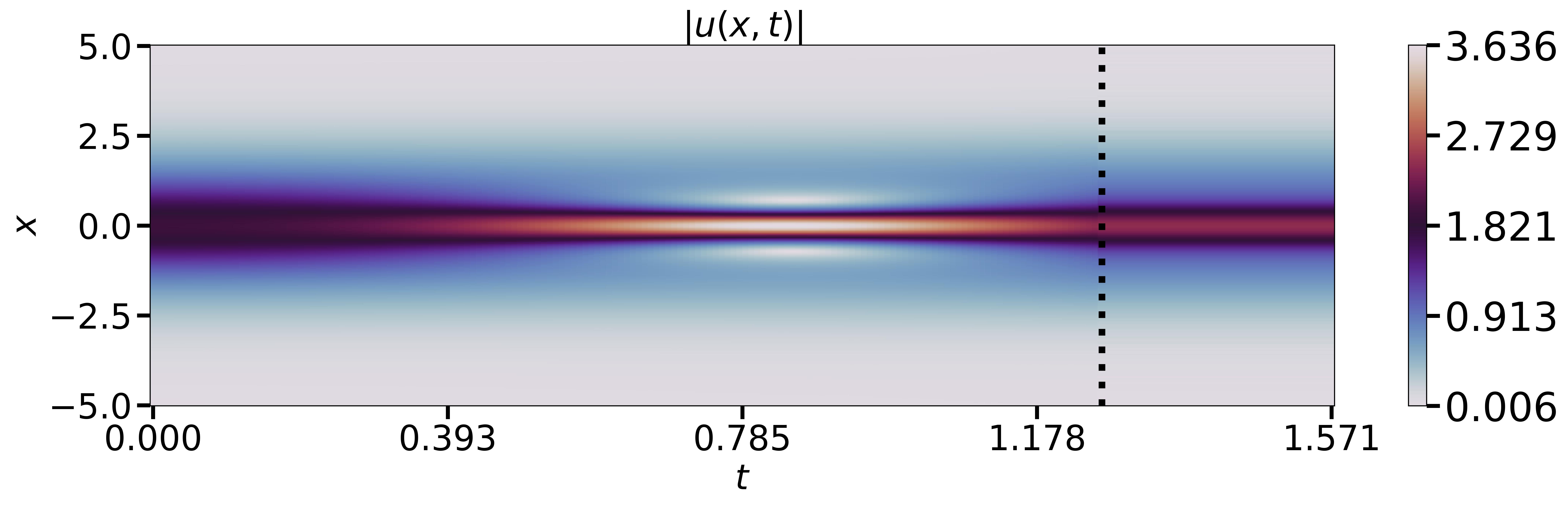}}\hfill
    \subfigure[CoRNN]{\label{fig:SC_CoRNN_contour}\includegraphics[width=0.23\textwidth]{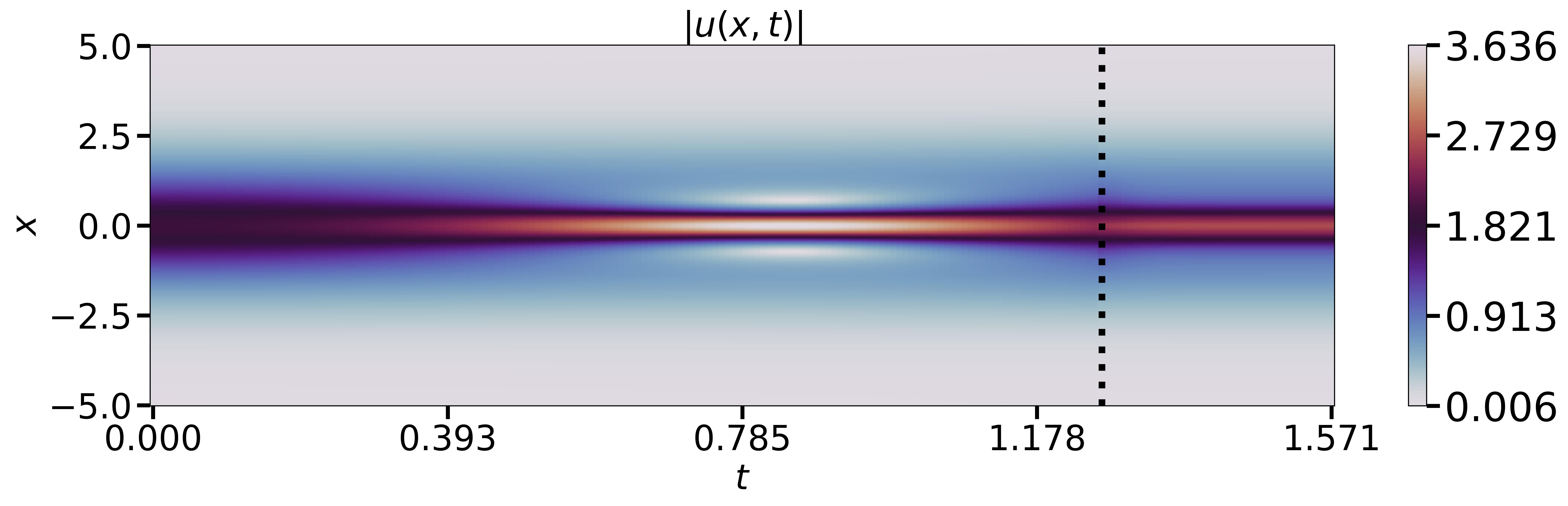}}\hfill
    \subfigure[LEM]{\label{fig:SC_LEM_contour}\includegraphics[width=0.23\textwidth]{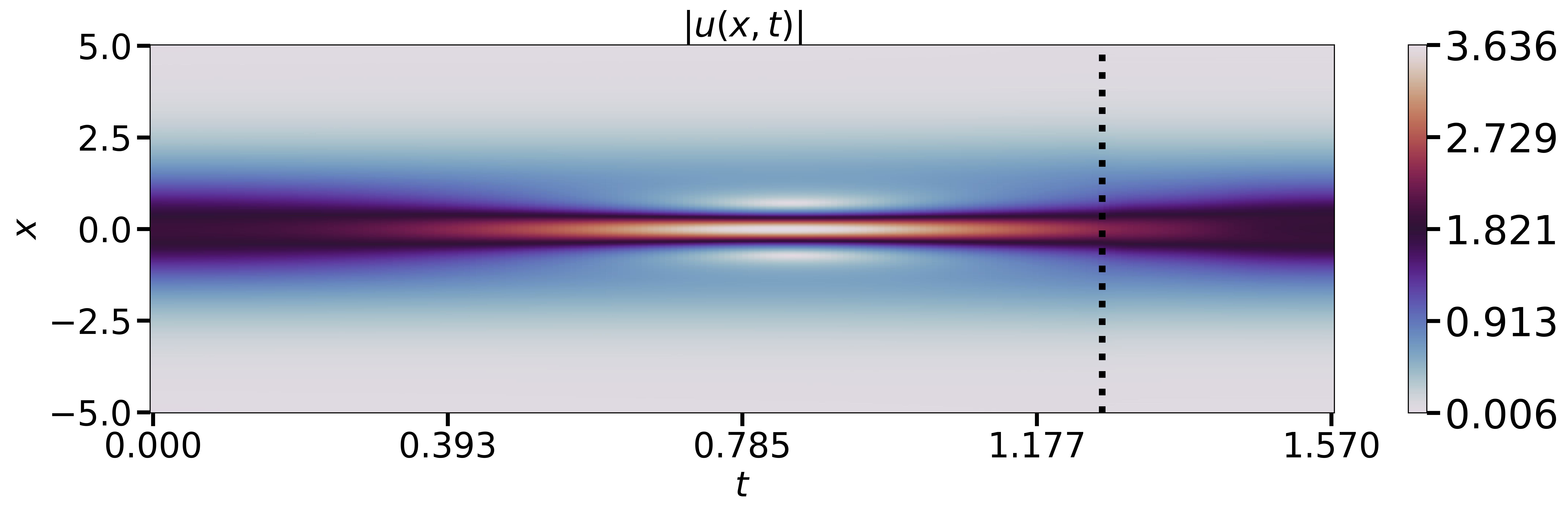}}\\
    \subfigure[GRU]{\label{fig4e}\includegraphics[height=2.5cm, width=2.5cm]{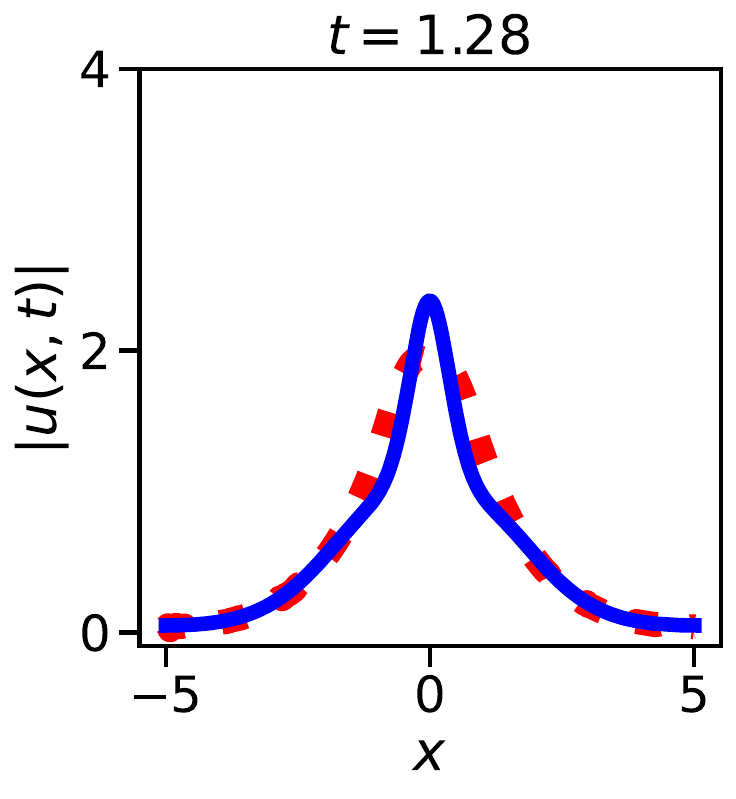}}\hfill
      \subfigure[CoRNN]{\label{fig4h}\includegraphics[height=2.5cm, width=2.5cm]{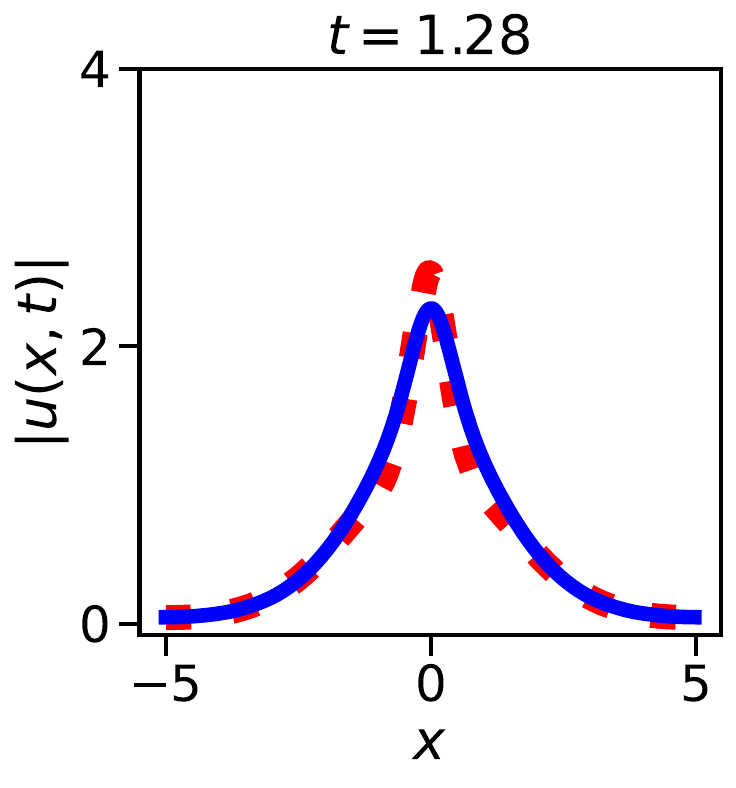}} \hfill
      \subfigure[LEM]{\label{fig4g}\includegraphics[height=2.5cm, width=2.5cm]{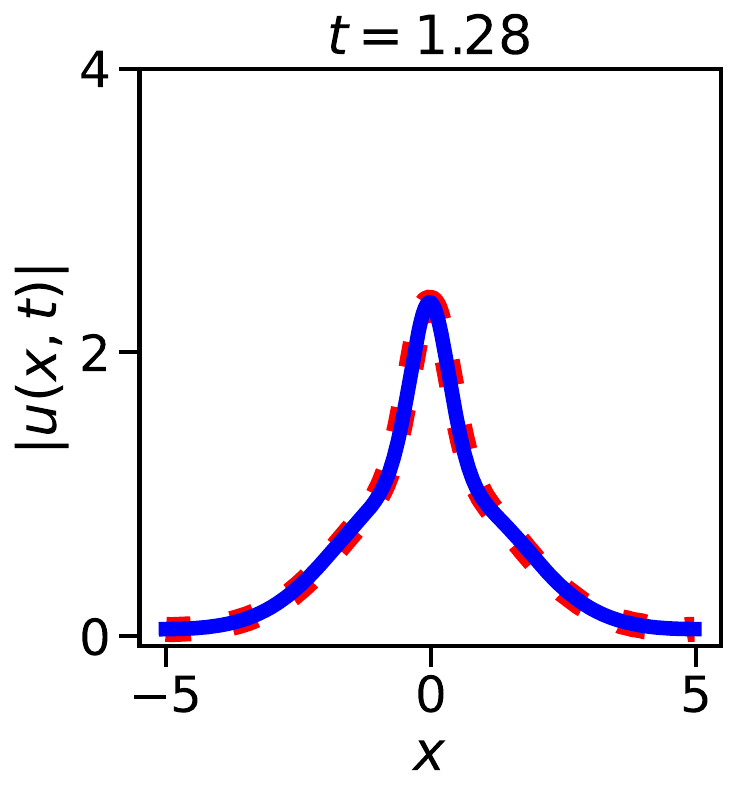}} \hfill
    \subfigure[GRU]{\includegraphics[height=2.5cm, width=2.5cm]{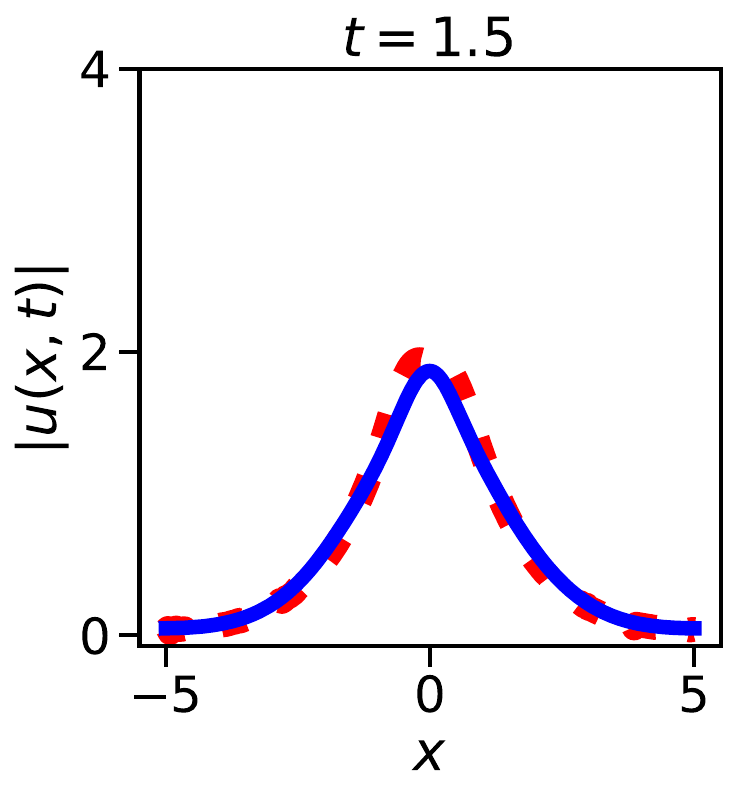}}\hfill
     \subfigure[CoRNN]{\includegraphics[height=2.5cm, width=2.5cm]{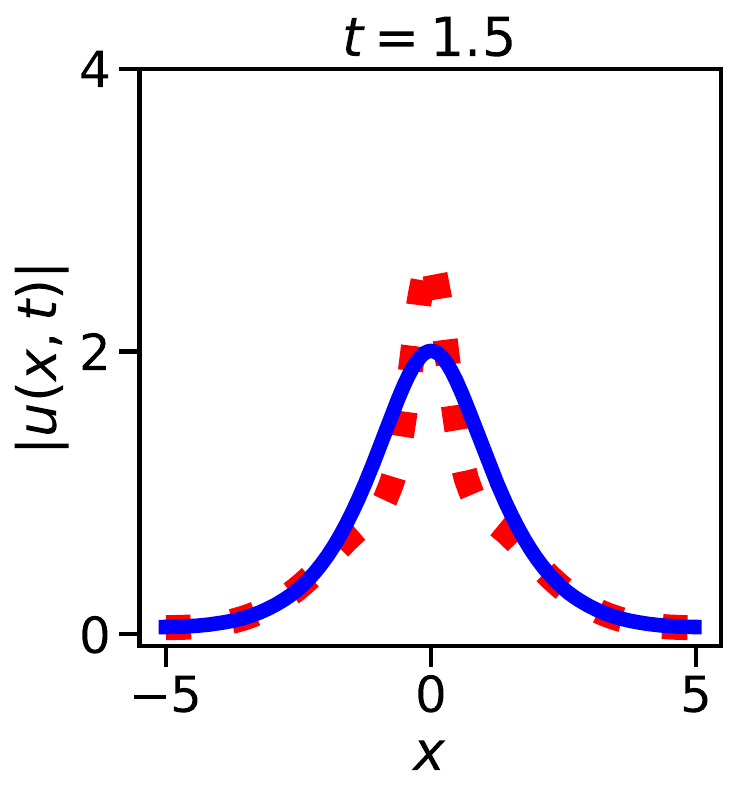}}\hfill
        \subfigure[LEM]{\label{fig4j}\includegraphics[height=2.5cm, width=2.5cm]{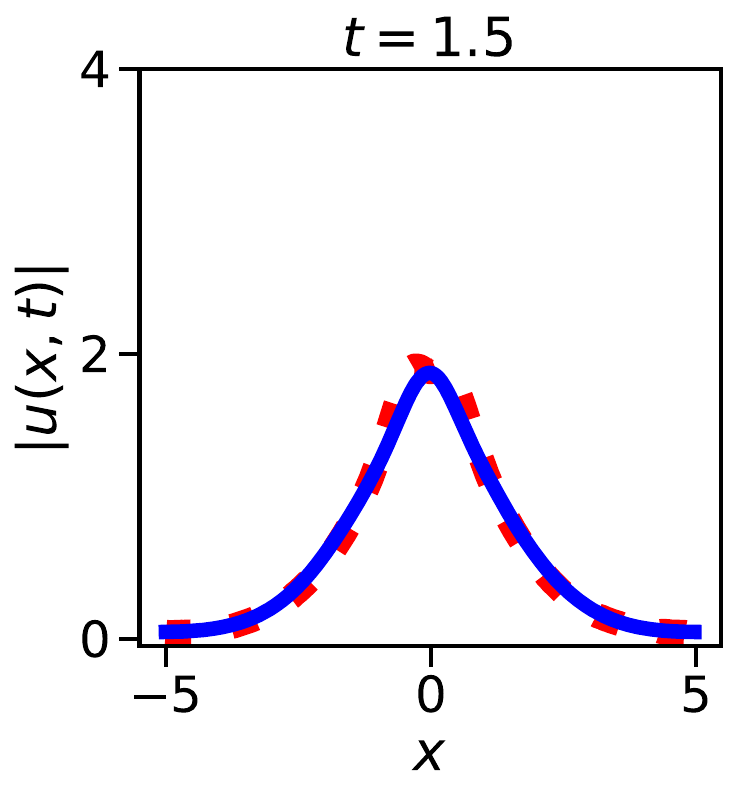}}\hfill
    \caption{Top two rows: the complete reference solution and predictions for the Schr\"{o}dinger equation. Bottom: the solution snapshots at $t=\{1.28, 1.5\}$ obtained in the generalization region.}
    \label{fig:SC}
\end{figure}

\begin{figure}[t]
    \centering
    \subfigure[Reference Solution]{\label{fig:EB_pinn_ref}\includegraphics[width=0.23\textwidth]{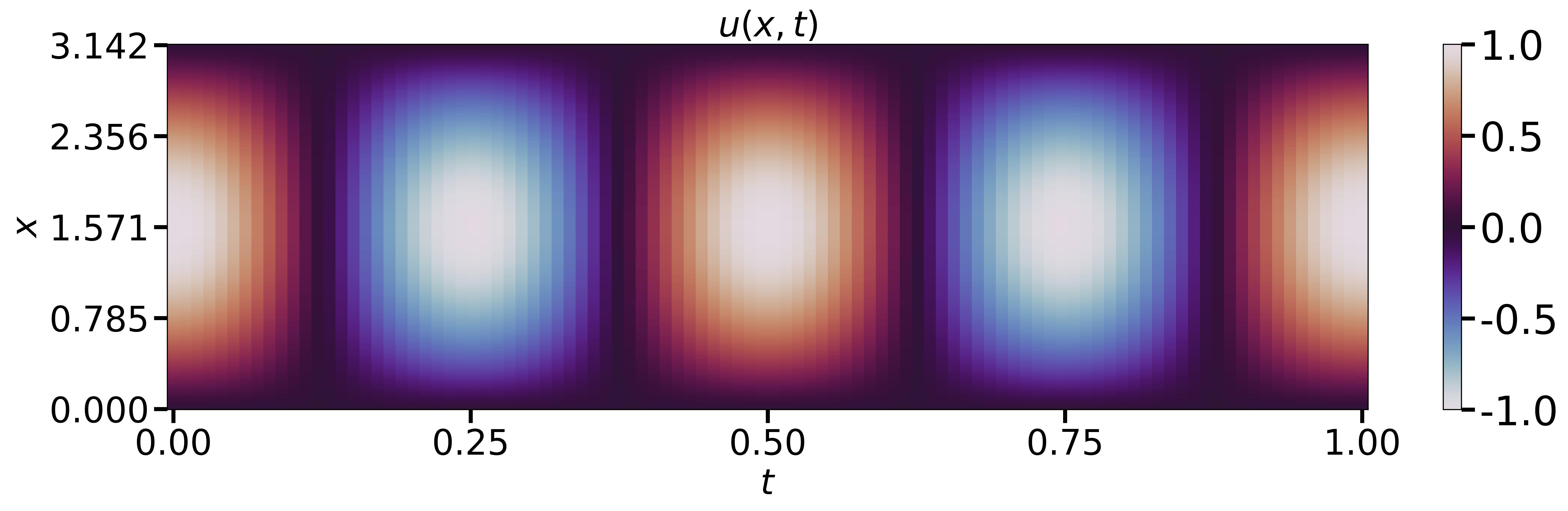}}\hfill
    \subfigure[GRU]{\label{fig:EB_GRU_contour}\includegraphics[width=0.23\textwidth]{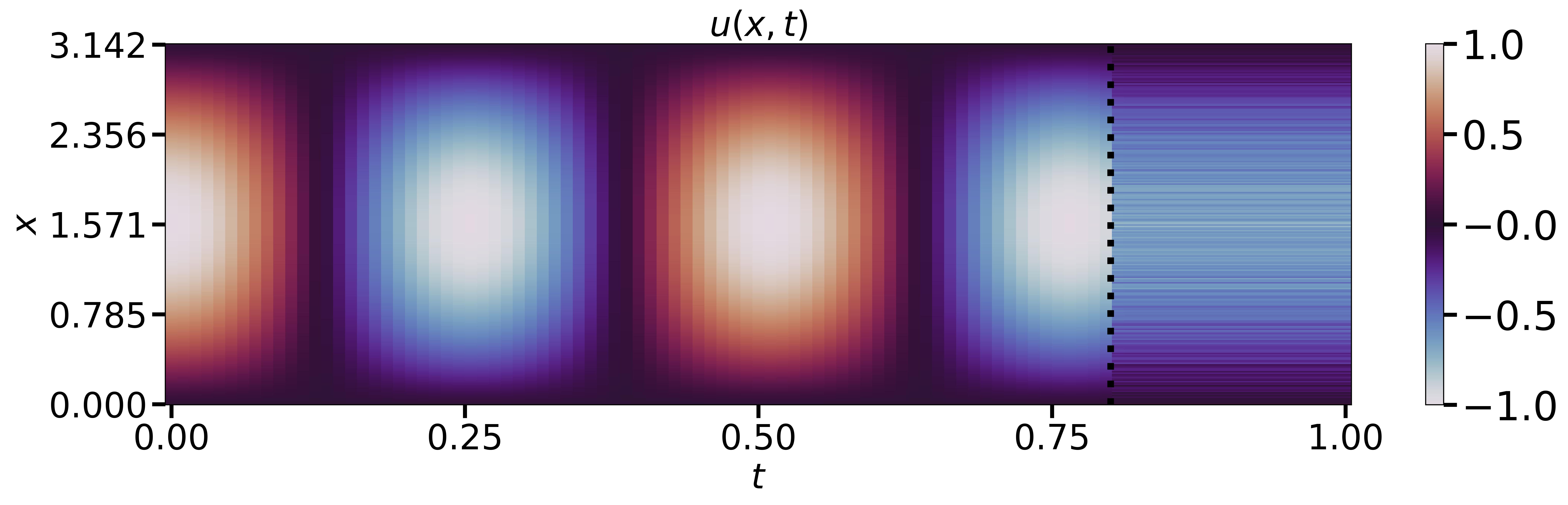}}\hfill
    \subfigure[CoRNN]{\label{fig:EB_CoRNN_contour}\includegraphics[width=0.23\textwidth]{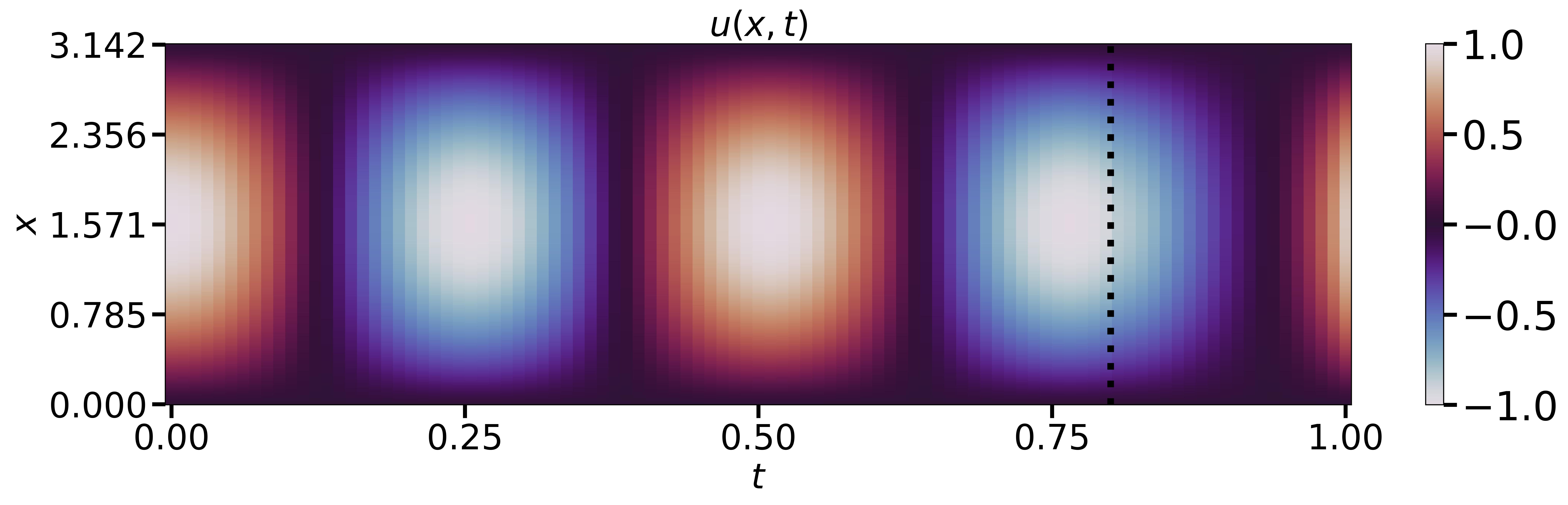}}\hfill
    \subfigure[LEM]{\label{fig:EB_LEM_contour}\includegraphics[width=0.23\textwidth]{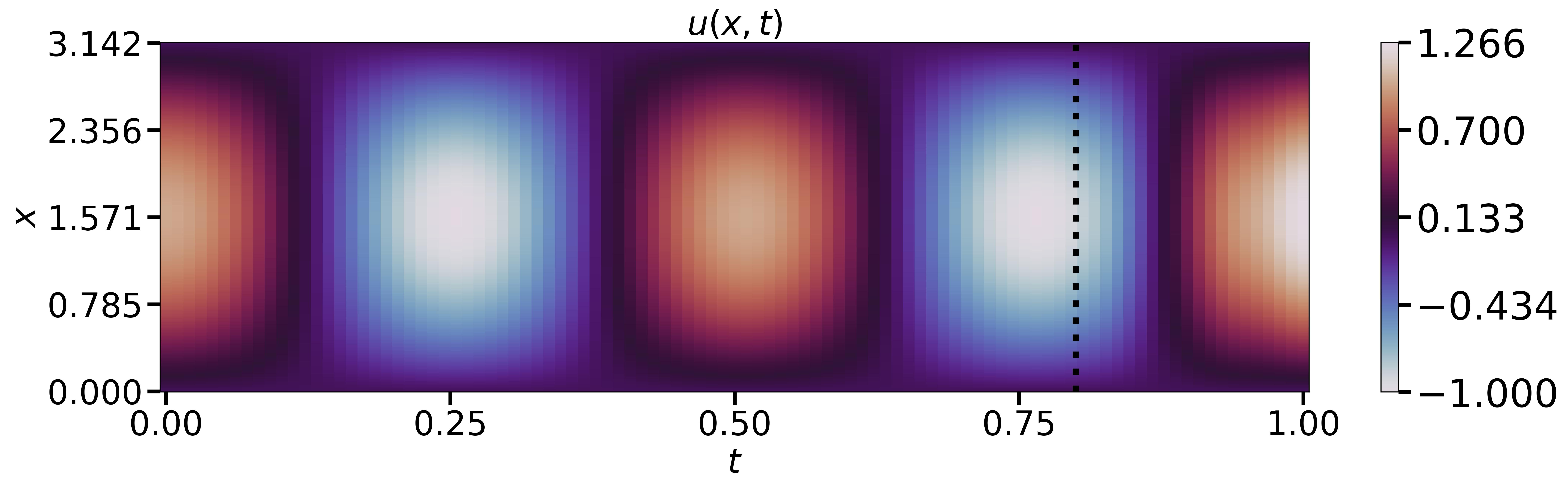}}\\
    \subfigure[GRU]{\label{fig5e}\includegraphics[height=2.5cm, width=2.5cm]{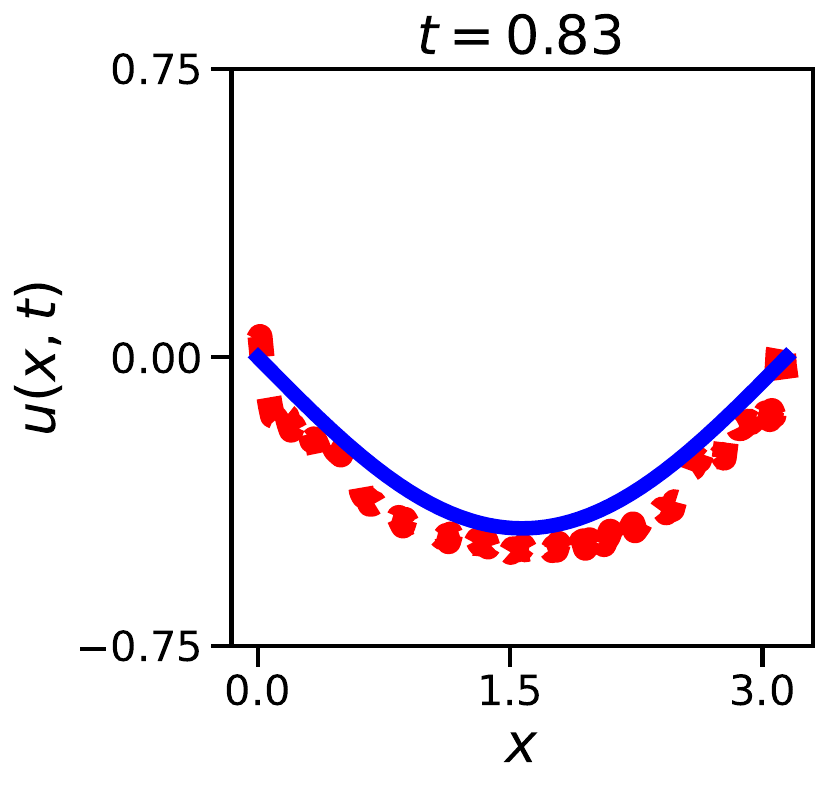}}\hfill
      \subfigure[CoRNN]{\includegraphics[height=2.5cm, width=2.5cm]{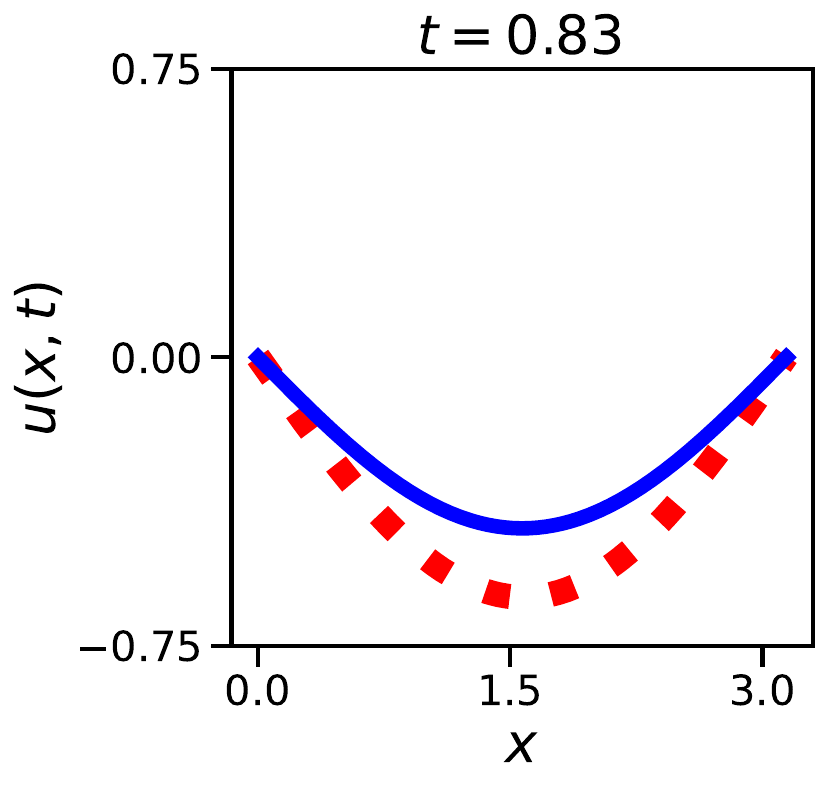}} \hfill
      \subfigure[LEM]{\label{fig5g}\includegraphics[height=2.5cm, width=2.5cm]{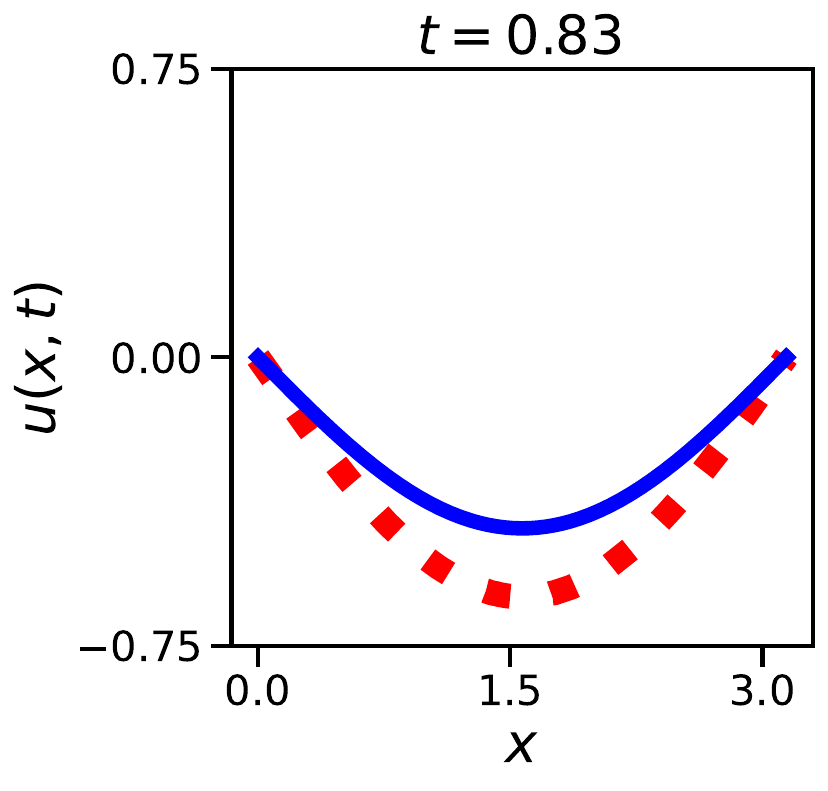}} \hfill
    \subfigure[GRU]{\label{fig5h}\includegraphics[height=2.5cm, width=2.5cm]{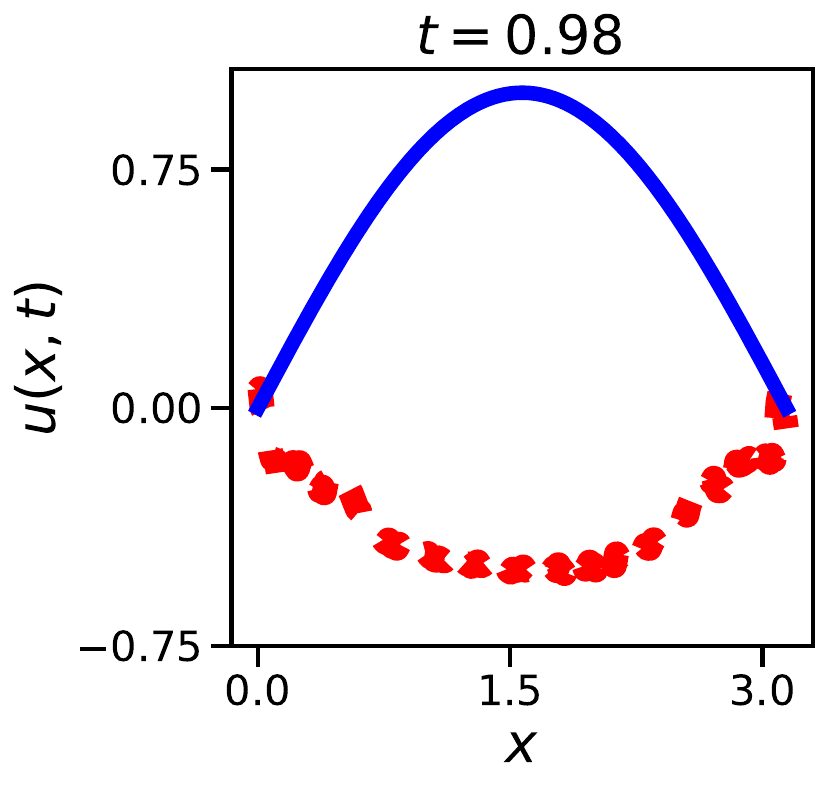}}\hfill
     \subfigure[CoRNN]{\label{fig5j}\includegraphics[height=2.5cm, width=2.5cm]{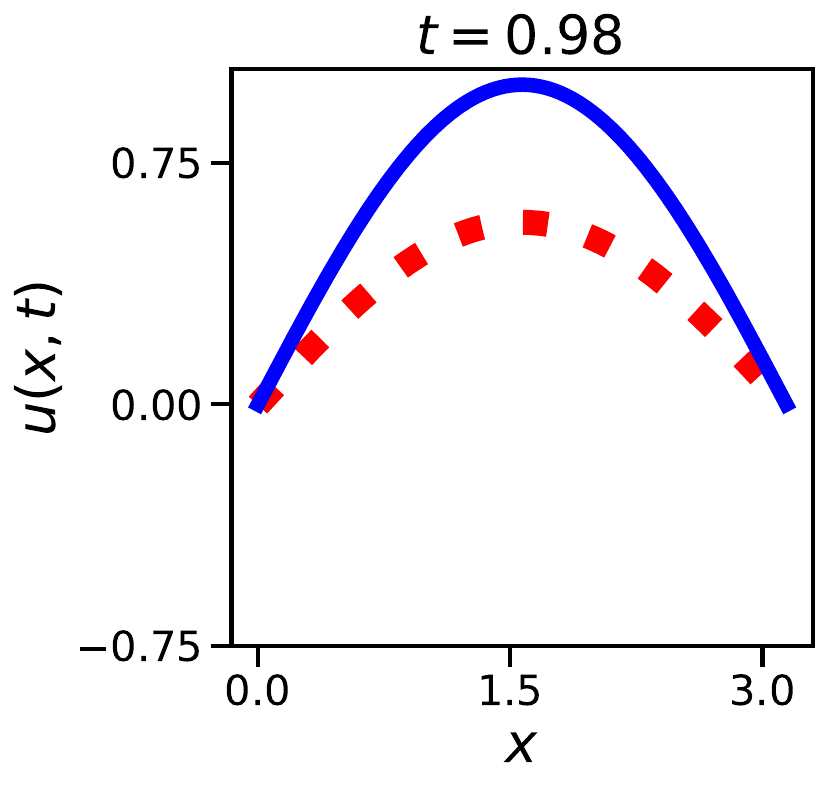}}\hfill
        \subfigure[LEM]{\includegraphics[height=2.5cm, width=2.5cm]{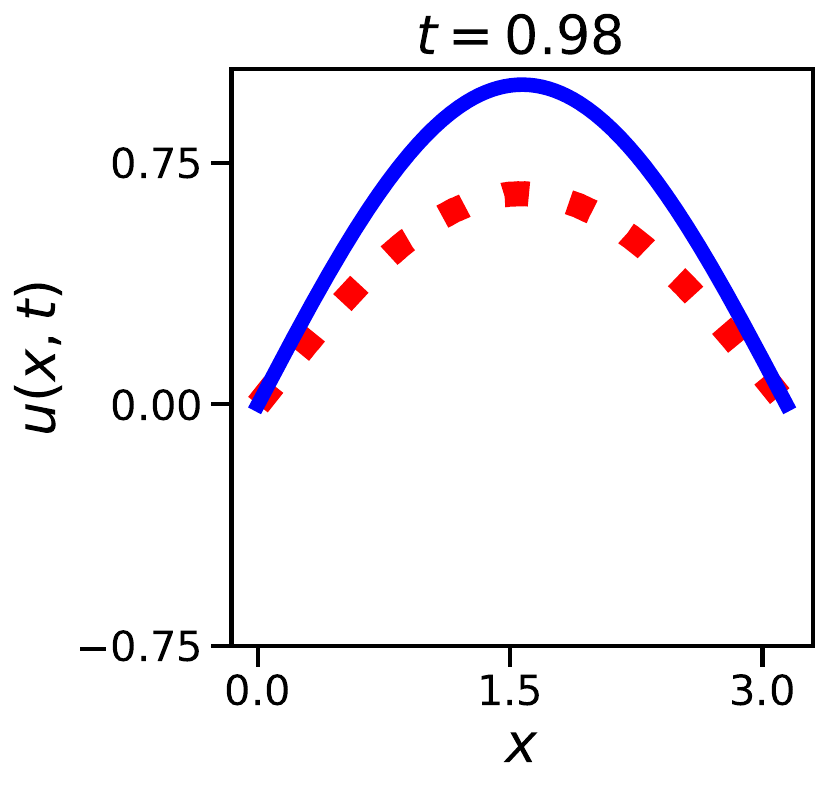}}\hfill
    \caption{Top two rows: the complete reference solution and predictions for the Euler--Bernoulli beam equation. Bottom: the solution snapshots at $t=\{0.83, 0.98\}$ obtained in the generalization region.} 
    \label{fig:EB}
\end{figure}

\section{Numerical Experiments}

We validate the proposed framework on three time-dependent nonlinear PDEs and a fourth-order biharmonic beam equation. The software and hardware environments used to perform the experiments are as follows: \textsc{Ubuntu} 20.04.6 LTS, \textsc{Python} 3.9.7, \textsc{Numpy} 1.20.3, \textsc{Scipy} 1.7.1, \textsc{Matplotlib} 3.4.3, \textsc{TensorFlow-gpu} 2.9.1, \textsc{PyTorch} 1.12.1, \textsc{CUDA} 11.7, and \textsc{NVIDIA} Driver 515.105.01, i7 CPU, and \textsc{NVIDIA GeForce RTX 3080}.

\subsubsection{PDEs} 

The four equations---viscous Burgers equation,  Allen-Cahn (AC) equation, nonlinear Schr\"{o}dinger equation (NLS) and Euler-Bernoulli beam equation---along with their boundary and initial conditions are provided in the supplementary material \textbf{SM}\S B. For training/testing, we divide the entire time domain into two segments: $T := [0, T_{\mathrm{train}}]$ and $T^{'} := (T_{\mathrm{train}}, T_{\mathrm{test}}]$, where $T_{\mathrm{test}} > T_{\mathrm{train}} > 0$. Our task is to predict the PDE solution in the convex testing hull \(X_2 = D \times T^{'}\) after the model has been trained on the convex training hull \(X_1 = D \times T\). For all the problems, $T_{\mathrm{train}} = 0.8  T_{\mathrm{test}}$, dividing the training and test sets in the ratio $4:1$, following the work of DPM \cite{kim2021dpm} to maintain uniformity. The domain for each PDE, i.e., $D, T$ and $T'$, is defined in \textbf{SM}\S B. 

\subsubsection{Baselines}
Our objective is to make predictions beyond \(X_1\), i.e., on \(X_2\), and to assess how well the trained models generalize. We compare the performance of PINNs with CoRNN or LEM on this task. We also compare our approach to the state-of-the-art DPM  \cite{kim2021dpm}. A comparative analysis is also carried out when traditional recurrent networks, RNN, LSTM, and GRU, are augmented with the physics-informed model instead of the oscillatory networks. 
This analysis provides insight into how well the oscillatory methods perform relative to traditional recurrent networks and gradient techniques when confronted with generalization tasks.

\subsubsection{Hyperparameters}

To predict a solution to Burgers equation in \(X_1\) using PINNs, $1600$ training points are used, comprising $1000$ residual points and $600$ points for boundary and initial time. The feedforward neural network has two inputs, space $x \in D$ and time $t \in T$. Four hidden layers, each containing 20 neurons, and hyperbolic tangent ($\tanh$) activation function are used to predict the approximation of the solution $u \in \mathcal{U}$. Optimization is performed using the $\textsc{LBFGS}$ algorithm for $3500$ epochs. For the Euler-Bernoulli beam equation, $16000$ training points are distributed as $10000$ residual points and $6000$ points designated for both initial and boundaries. The hyperparameters are kept the same as in the viscous Burgers equation. Allen-Cahn and Schr\"{o}dinger equations are simulated using the software DeepXDE \cite{lu2021deepxde} with the default hyperparameters described therein.
 
The input and output size of the recurrent networks is taken to be $k_x$, with a single hidden layer of size $32$. The sequence length is chosen to be $k_t$. The exact values of $k_x$ and $k_t$ are defined in the ``Train and test criteria" subsection for each equation. $\textsc{Adam}$ optimizer is used to train the recurrent networks. The learning rates for LEM, CoRNN, GRU, LSTM, and RNN are $0.001$, $0.001$, $0.01$, $0.01$, and $0.01$, respectively, across all equations. For Schr\"{o}dinger equation, a learning rate of $0.01$ is used to train the LEM. In the case of CoRNN, two additional hyperparameters, $\gamma$ and $\epsilon$, are set to $1.0$ and $0.01$, respectively. The number of epochs executed for Burgers and Allen--Cahn equations is $20,000$, while for Schr\"{o}dinger equation, it is $30,000$. Lastly, $200,000$ epochs are performed for the Euler-Bernoulli beam equation.

\subsubsection{Evaluation metrics}
For the first three experiments, the errors are reported relative to the numerical solutions of the corresponding PDEs. 
The reference for the Euler-Bernoulli beam equation is an analytical solution described in \textbf{SM}\S B. As the criteria for assessment, we employ standard evaluation metrics: the relative errors in the L2-norm, the explained variance score, the maximum error, and the mean absolute error, defined in \textbf{SM}\S D. Each of these metrics provides distinct insights into the performance. Furthermore, we present visual snapshots of both the reference and approximate solutions at specific time instances. Additional snapshots and contour results are provided in \textbf{SM}\S C. 

\subsubsection{Train and test criteria}

The trained PINN is tested on $k_t \cdot k_x$ points in \(X_1\). For the Burgers equation and the Euler-Bernoulli beam equations, we set $k_x = 256$ and $k_t = 80$. For the Allen-Cahn equation, $k_x = 201$ and $k_t = 80$. For the Schr\"{o}dinger equation, $k_x = 256$ and $k_t = 160$.

The PINN output provides input to train the neural oscillators, adhering to the specified hyperparameter configuration. After training the neural oscillator on \(X_1\), testing is extended to \(X_2\). This testing sequence commences at $\inf(T{'})$ as the initial input. The ensuing output is then utilized as the input for the subsequent sequence (Fig.~\ref{fig:proposed_architecture}). Such testing is crucial since, in practical scenarios, knowledge about the solution $u$ in \(X_2\) is absent. Thus, the solely available information for generalization is derived from the predicted solution within \(X_2\). This testing process is iterated until reaching $\sup(T')$. The domains \(X_1\), \(X_2\) and $T^{'}$ for all the equations are provided in \textbf{SM}\S B.

\subsection{Experimental Results}

\begin{table*}
\setlength{\tabcolsep}{2pt}
\centering
\scriptsize 
\caption{The generalization accuracy in terms of the relative errors in the L2-norm, the explained variance error, the max error, and the mean absolute error for various PDEs. Higher (or lower) values are preferred, corresponding to $\uparrow$ (or $\downarrow$).}\label{tbl:2}
\begin{tabular}{|c|c|c|c|c|c|c|c|c|c|c|c|c|c|c|c|c|}
\hline
\multirow{2}{*}{PDE} & \multicolumn{4}{c|}{L2-norm $(\downarrow)$} & \multicolumn{4}{c|}{Explained variance score $(\uparrow)$} & \multicolumn{4}{c|}{ Max error $(\downarrow)$} & \multicolumn{4}{c|}{ Mean absolute error $(\downarrow)$} \\ \cline{2-17}
          &RNN & LSTM & GRU   & LEM & RNN & LSTM & GRU & LEM & RNN & LSTM & GRU & LEM & RNN & LSTM & GRU&LEM  \\ \hline
Vis. Burgers & 0.4154 & 0.4635 & 0.3768 &\textbf{0.0001} & 0.5845  & 0.5364  & 0.6231 & \textbf{0.9998}  & 0.5943 & 1.0403 & 0.5447 &\textbf{0.0246}  & 0.2662&0.2856 & 0.2530&\textbf{0.0035} \\ \hline
Allen--Cahn &  0.0058 & 0.0570 &0.0093  & \textbf{0.0049} & 0.9951  & 0.9469 & 0.9919 &\textbf{0.9956}  & 0.1457 & 0.3996 & 0.1763 & \textbf{0.1376} &0.0406 & 0.1946 & 0.0508 &\textbf{0.0348}  \\ \hline
Schr\"{o}dinger &0.3170   &0.5022  & 0.0218  & \textbf{0.0034}  & 0.4408 & 0.2721 &0.9619  & \textbf{0.9944} & 1.6950 & 0.1532 & 0.4347 & \textbf{0.0948} &0.2601  &0.3268   & 0.0756 & \textbf{0.0281} \\ \hline
Euler--Bernoulli & 4.6509  &2.1198  & 2.9176  & \textbf{0.0593}  & -0.8447 & -0.2583 & -0.5666 & \textbf{0.9409} & 1.9976 &1.4652  &2.0449  & \textbf{0.2673} & 0.7976 & 0.5000  & 0.6046 &\textbf{0.0915}  \\ \hline
\end{tabular}
\end{table*}

Tables~\ref{tbl:1} and~\ref{tbl:2} collate the overall performance metrics for the oscillator-based methods (LEM, CoRNN) in comparison with DPM, RNN, LSTM and GRU. The results show that LEM exhibits significantly superior performance across all the benchmark problems.

\subsubsection{Viscous Burgers equation} 

Figure~\ref{fig:burgers} provides a visual comparison between the reference solution (Fig.~\ref{fig:visc_pinn_ref}) and its counterparts generated with GRU, CoRNN and LEM (Figs.~\ref{fig:Burger_GRU_contour}--\ref{fig:Burger_LEM_contour}, respectively). GRU struggles to accurately capture the solution of  Burgers equation, leading to the loss in prediction accuracy as time \(t\) increases. Our methods based on CoRNN and LEM exhibit notably improved predictive accuracy, even when \(t\) approaches the end of the time domain. Figures~\ref{fig2e}--\ref{fig2j} provide further insights into the solution at time instances $t = 0.83, 0.98$. They reveal that LEM outperforms the alternative methods across the entire space-time domain. The performance of CoRNN is comparable to that of LEM, producing reasonably accurate predictions. These findings underscore the significance of neural oscillators in precise generalization. Additional experiments on \emph{sensitivity analysis} of oscillator parameter ($\Delta t$) along with an \emph{ablation study} on CoRNN parameters $\epsilon$ and $\gamma$ is presented in \textbf{SM}\S C. Additionally, the generalization in parametric space \cite{kapoor2023neural} is also presented in \textbf{SM}\S C.

\subsubsection{Allen-Cahn equation} 

In Figure~\ref{fig:AC}, the reference solution of the Allen-Cahn equation is compared to its counterparts generated with GRU, CoRNN and LEM. Our oscillator-based methods (CoRNN and LEM) yield the most precise approximations in the generalization domain (Figs.~\ref{fig:AC_pinn_ref}--\ref{fig:AC_LEM_contour}). 
The LEM-based solution exhibits a nearly symmetric behavior with respect to \(x = 0\), demonstrating its ability to preserve the symmetry and structure of the solution. At \(t = 0.81\), all three methods display a similar level of accuracy (Figs.~\ref{fig3e}--\ref{fig3g}). However, as time advances, e.g., at \(t = 0.99\), the performance of LEM surpasses that of the other techniques throughout the extrapolation domain (Figs.~\ref{fig3h}--\ref{fig3j}).

\subsubsection{Schr\"{o}dinger equation} 

Figure~\ref{fig:SC} illustrates a comparison between the reference solution of Schr\"{o}dinger equation and its counterparts generated with GRU, CoRNN and LEM. Rather than plotting the real and imaginary parts of this solution, Figs.~\ref{fig:SC_pinn_ref}--\ref{fig:SC_LEM_contour} exhibit its magnitude, \(|u(x, t)|\); the solutions are visually indistinguishable. The three approximations are accurate at time $t=1.28$ (Figs.~\ref{fig4e}--\ref{fig4g}), but the GRU- and CoRNN-based solutions at \(t = 1.5\) have errors around \(x = 0\) whereas the LEM-based solution retains its accuracy within that region (Figs.~\ref{fig4h}--\ref{fig4j}). 

\subsubsection{Euler-Bernoulli beam equation} 

In Figure~\ref{fig:EB}, we compare the analytical solution of the  Euler-Bernoulli beam equation to approximate solutions obtained with GRU, CoRNN and LEM. The intricacy of this linear equation stems from the presence of fourth-order derivatives \cite{kapoor2023physics, cao2023lno}, rendering it a compelling challenge for the proposed methodology . The visual comparison afforded by Figs.~\ref{fig:EB_pinn_ref}--\ref{fig:EB_LEM_contour} demonstrates the superiority of the LEM-based solution and the inferiority of the GRU-based one. At \(t = 0.83\), all three approximations are qualitatively correct, with various degrees of accuracy (Figs.~\ref{fig5e}--\ref{fig5h}). At \(t = 0.98\), the GRU-based solution is not only inaccurate but is also qualitatively incorrect, while the oscillator-based approximators correctly predict the system's behavior (Figs.~\ref{fig5h}--\ref{fig5j}). 

\section{Conclusion}

We introduced a method that combines neural oscillators with physics-informed neural networks to enhance performance in unexplored regions. This novel approach enables the model to learn the long-time dynamics of solutions to the governing partial differential equations. We demonstrated the effectiveness of our method on three benchmark nonlinear PDEs: viscous Burgers, Allen-Cahn, and Schr\"{o}dinger equations, as well as the biharmonic Euler-Bernoulli beam equation. Our results showcase the improved generalization performance of the PIML augmented with neural oscillators, which outperforms state-of-the-art methods in various metrics. The codes to reproduce the presented results are provided at https://github.com/taniyakapoor/AAAI24\_Generalization\\\_PIML.

\bibliography{aaai24}

\clearpage
\appendix
\section{Supplementary Material} 
\section{A. Nomenclature}

The table provided below presents the abbreviations utilized within this paper.

\begin{table}[h]
\caption{Abbreviations used in this paper}
\vskip 0.15in
\begin{center}
\begin{small}
\begin{sc}
\begin{tabular}{ll}
\toprule
 Symbol & Description    \\
\midrule
\text{AC} & \text{Allen--Cahn} \\
\text{CoRNN} & \text{Coupled oscillatory recurrent neural network} \\
\text{DPM} & \text{Dynamic pulling method} \\
\text{EVGP} & \text{Exploding and vanishing gradient problem} \\
\text{GRU} & \text{Gated recurrent unit} \\
\text{LSTM} & \text{Long short-term memory} \\
\text{MAE} & \text{Mean absolute error} \\
\text{NLS} & \text{Nonlinear Schr\"{o}dinger equation} \\
\text{ODE} & \text{Ordinary differential equation} \\
\text{PDE} & \text{Partial differential equation} \\
\text{PIML} & \text{Physics-informed machine learning} \\
\text{PINN} & \text{Physics-informed neural network} \\
\text{RNN} & \text{Recurrent neural network} \\
\text{RMSE} & \text{Root mean squared error} \\
\text{SM} & \text{Supplementary material} \\
\text{SOTA} & \text{State-of-the-art} \\
\bottomrule
\end{tabular}
\end{sc}
\end{small}
\end{center}
\vskip -0.1in
\end{table}

\section{B. PDEs: Domains and Conditions}
In the following subsections, PDEs for the considered problems are presented, accompanied by their respective domains as well as initial and boundary conditions. For all the PDEs, \(X_1 := D \times T\), and \(X_2 := D \times T^{'}\).

\subsection{Viscous Burgers equation}

\begin{gather}
u_\mathrm{t} + uu_\mathrm{x} - (0.01/\pi) u_{xx}  = 0, \\ x\in D := [-1, 1]; \quad t \in T := [0, 0.8]; \quad T' := (0.8, 1]
    \label{eqBur}
\end{gather}

with initial and boundary conditions
\begin{equation*}
u(x, 0) = -\sin(\pi x)
\end{equation*}
\begin{equation*}
    u(-1, t) = u(1, t)  = 0
\end{equation*}

\subsection{Allen--Cahn equation}

\begin{gather}
    u_\mathrm{t} - 0.0001 u_{xx} + 5u^3 - 5u = 0, \\ x\in D := [-1, 1]; \quad t \in T := [0, 0.8]; \quad T' := (0.8, 1]
    \label{eqAC}
\end{gather}

with initial and periodic boundary conditions

\begin{equation*}
u(x, 0) = x^2 \cos(\pi x) \text{sech}(x)
\end{equation*}
\begin{equation*}
    u(-1, t) = u(1, t); \quad u_\mathrm{x}(-1, t) = u_\mathrm{x}(1, t) 
\end{equation*}

\subsection{Nonlinear Schr\"{o}dinger equation}

\begin{gather}
    u_\mathrm{t} - i0.5 u_{xx} - i |u|^2u = 0, \\ x\in D := [-5, 5]; \quad t \in T := [0, 2 \pi /5]; \quad T' := ( 2 \pi /5, \pi/2]
    \label{eqSC}
\end{gather}

with initial and periodic boundary conditions

\begin{equation*}
u(x, 0) = 2 \text{sech}(x)
\end{equation*}
\begin{equation*}
    u(-5, t) = u(5, t); \quad u_\mathrm{x}(-5, t) = u_\mathrm{x}(5, t)
\end{equation*}

\subsection{Euler--Bernoulli beam equation}
\begin{gather}
    u_\mathrm{tt} + u_\mathrm{xxxx} = f(x, t), \\ x\in D := [0, \pi]; \quad t \in T := [0, 0.8]; \quad T' := (0.8, 1]
    \label{eqEB}
\end{gather}

where $f(x, t) = (1 - 16\pi^2)\sin{(x)}\cos(4\pi t)$. The initial and boundary conditions are

\begin{equation*}
u(x, 0) = \sin(x), \quad u_{t}(x, 0) = 0
\end{equation*}
\begin{equation*}
    u(0, t) = u(\pi, t) = u_\mathrm{xx}(0, t) = u_\mathrm{xx}(\pi, t) = 0
\end{equation*}

The analytical solution for this problem is

\begin{equation*}
    u(x, t) =  \sin(x)\cos(4\pi t)
\end{equation*}

\section{C. Additional Results}
The following subsections present the additional obtained results for RNN and LSTM for various equations.

\subsection{Viscous Burgers equation}
Fig. 6 presents the contour plot of the approximations of the solution for the Burgers equation, along with snapshots at specific time instances (t = 0.83 and 0.98) for RNN and LSTM models.

\begin{figure}[H]
    \centering
    \subfigure[RNN]{\label{fig:AC_pinn_ref_supp1}\includegraphics[width=0.23\textwidth]{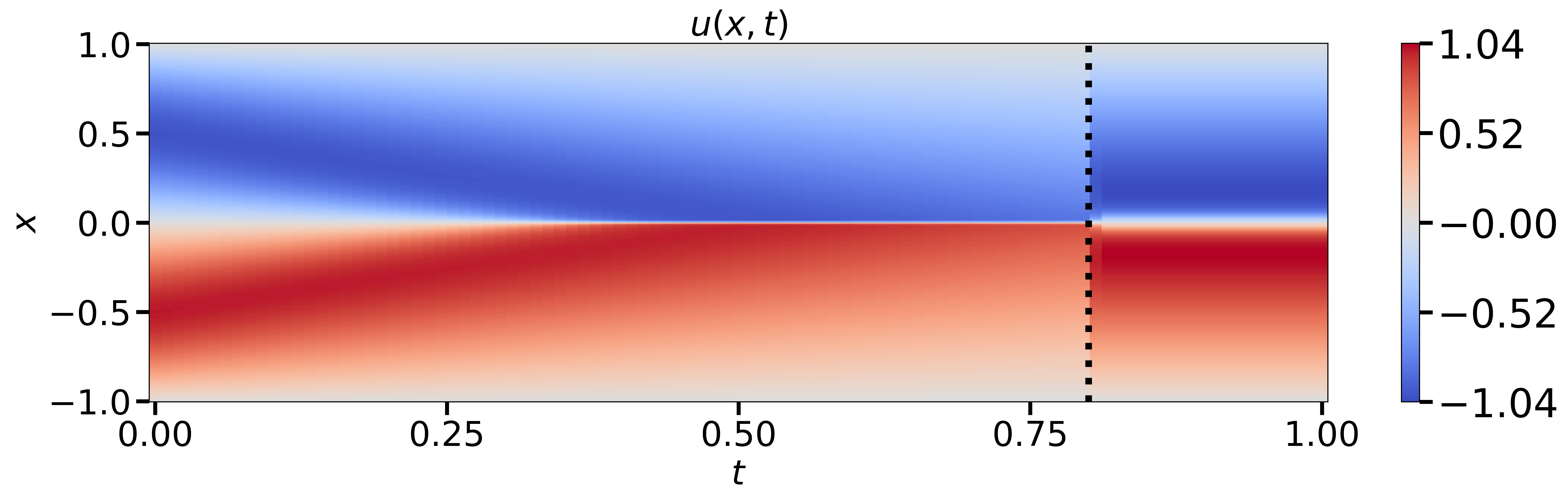}}\hfill
    \subfigure[LSTM]{\label{fig:AC_GRU_contour_supp1}\includegraphics[width=0.23\textwidth]{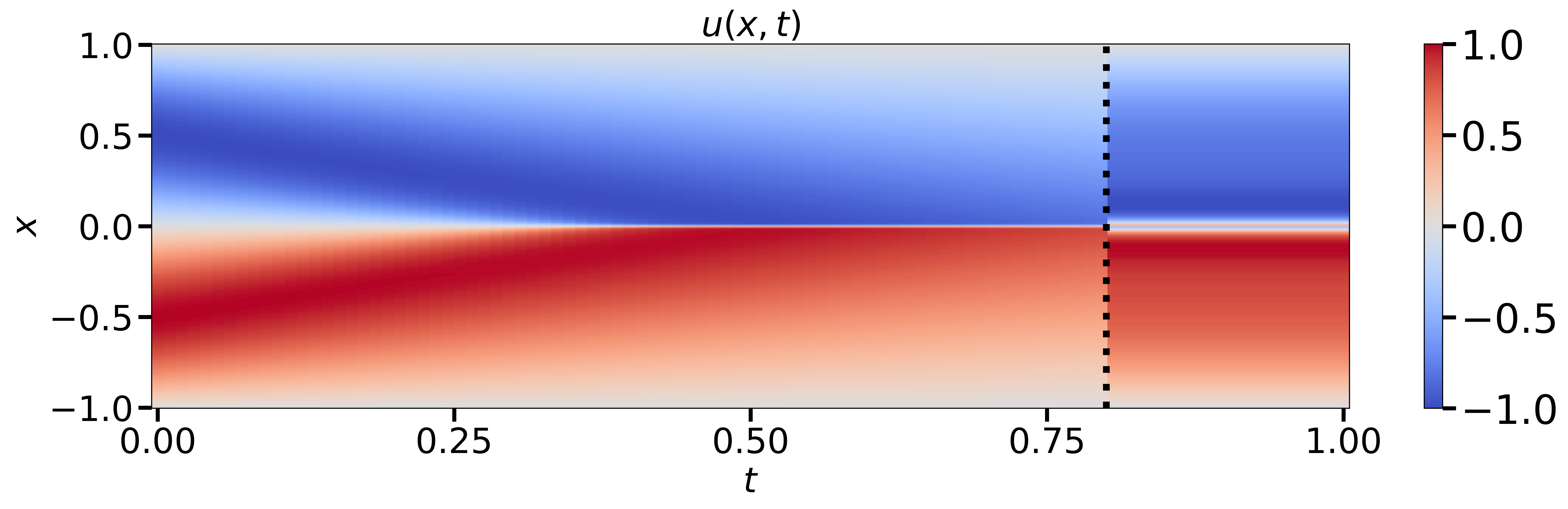}}\hfill
    \subfigure[RNN]{\includegraphics[height=3.5cm, width=3.5cm]{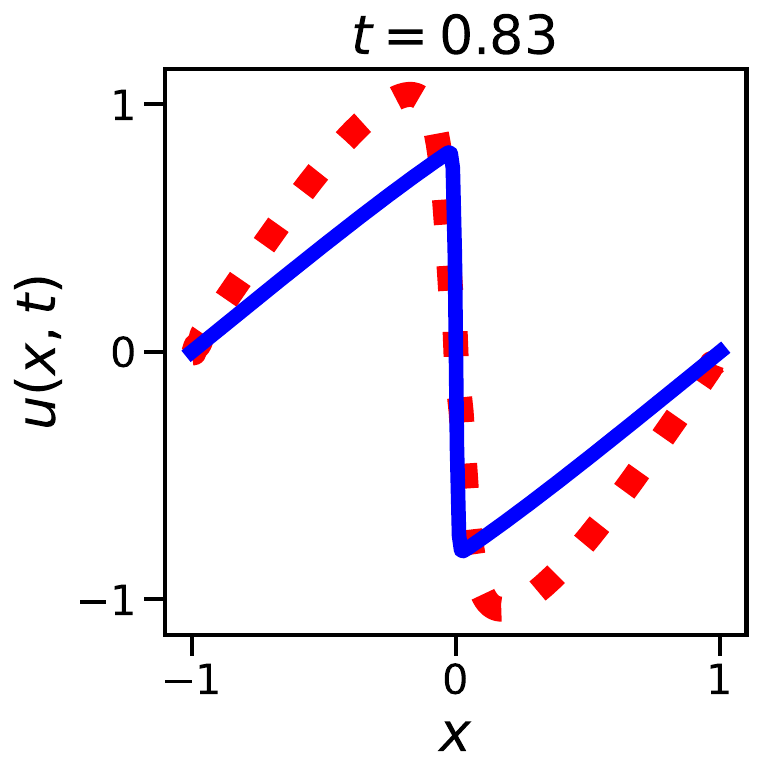}}\hfill
      \subfigure[LSTM]{\includegraphics[height=3.5cm, width=3.5cm]{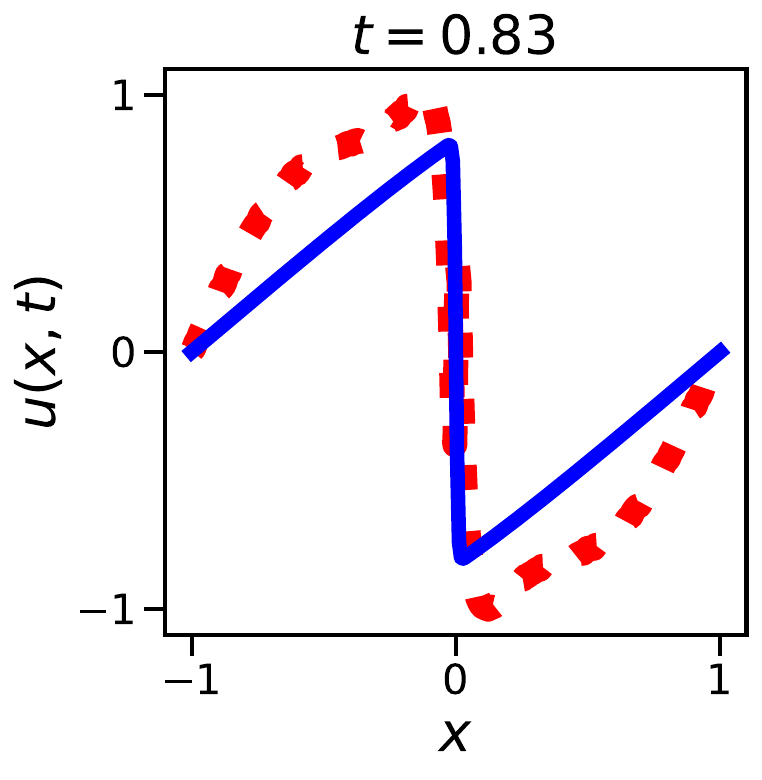}} \hfill
      \subfigure[RNN]{\includegraphics[height=3.5cm, width=3.5cm]{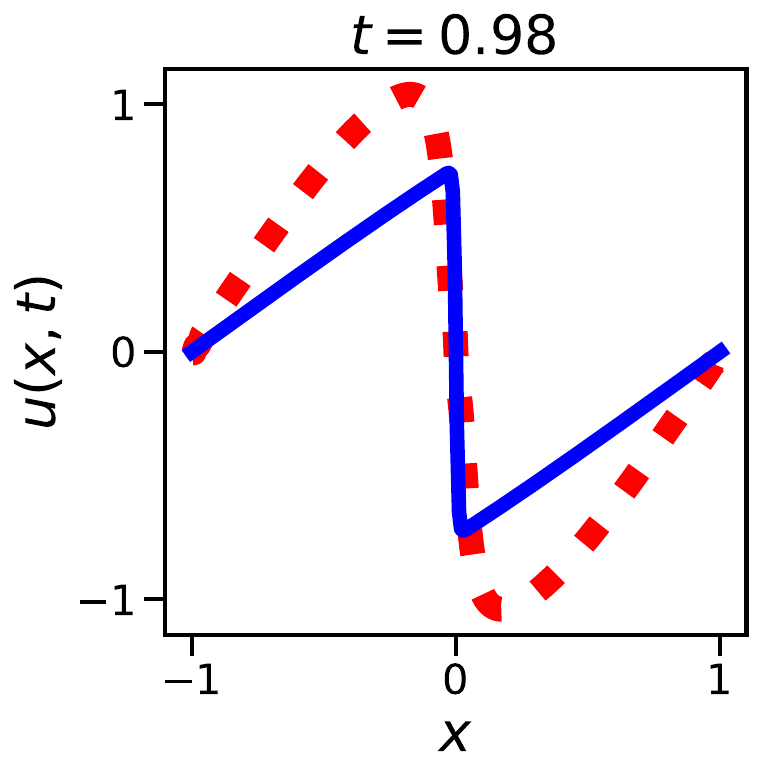}} \hfill
    \subfigure[LSTM]{\includegraphics[height=3.5cm, width=3.5cm]{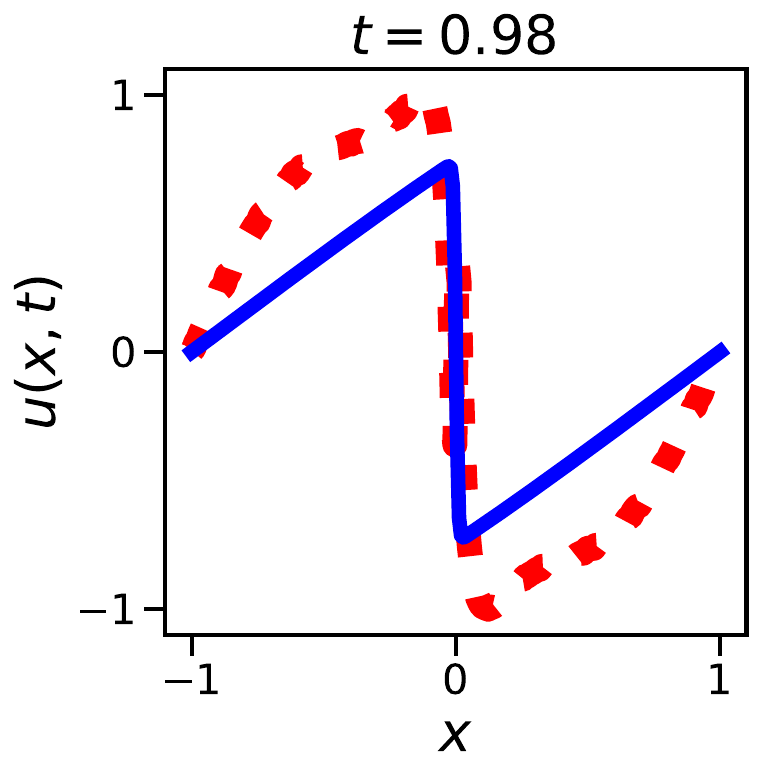}}\hfill
    \caption{Top row: predictions for the Burger equation for RNN and LSTM. Bottom row: the solution snapshots at $t=\{0.83, 0.98\}$ obtained in the generalization region.}
\end{figure}

We observe that the overall performance of the pipeline depends on the accuracy of PINN. However, the dependence is less than linear, as seen from the relative $L^2$ error table below for the Burgers equation.\\

\begin{tabular}{|c|c|c|c|c|}
\hline
PINN & 0.00026 & 0.02649 & 0.66240 & 2.64963 \\ \hline
Overall & 0.00792 & 0.04912 & 1.18762 & 4.74776 \\
\hline
\end{tabular}

\subsubsection{Sensitivity Analysis, Ablation study and Generalization in Parametric Space}
We perform an additional experiment for Burgers equation on \emph{sensitivity analysis} of oscillator parameter ($\Delta t$), varied for five different values in [0.1, 0.9]. Each experiment is run five times, and the mean and std dev is shown for LEM and CoRNN in Fig. 7 (\textit{left}), showcasing lower $\Delta t$ provides lower errors.
\begin{figure}[H]
    \centering
    \begin{minipage}[b]{0.49\textwidth}
        \centering
        \includegraphics[width=0.33\textwidth]{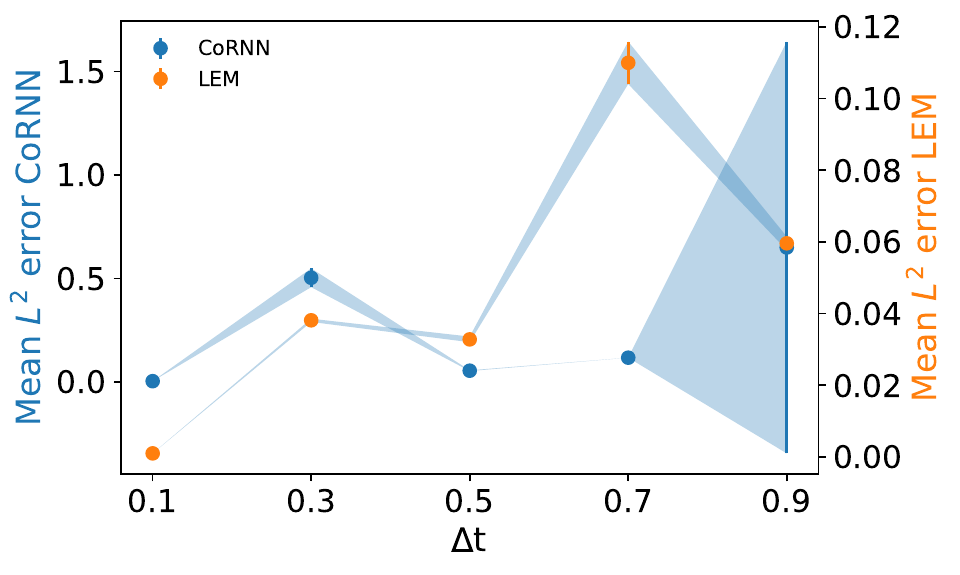}
         \includegraphics[width=0.32\textwidth]{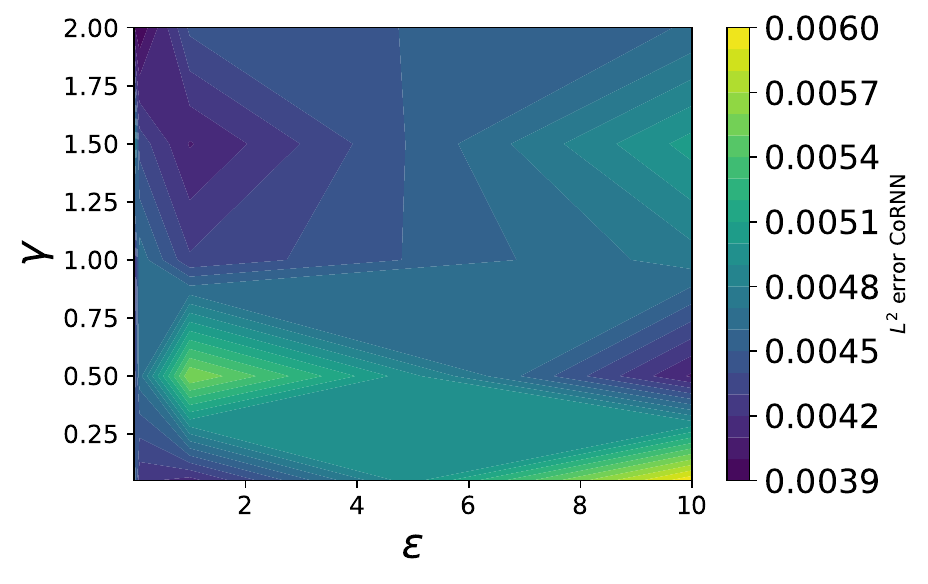}
         \includegraphics[width=0.32\textwidth]{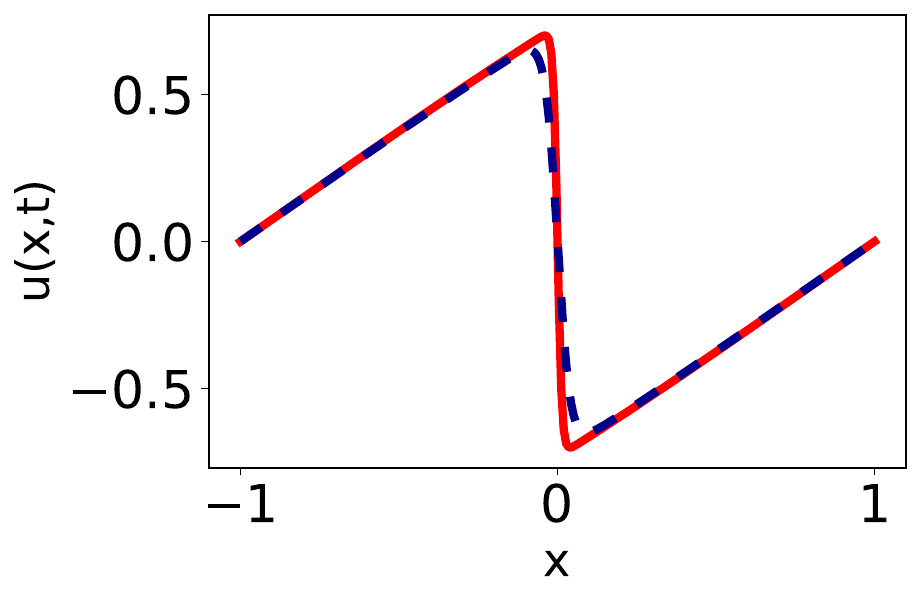}
    \end{minipage}
    \caption{Additional experiments on sensitivity analysis (\textit{left}), ablation study (\textit{middle}), and parametric space (\textit{right}).}
    \label{fig1}
\end{figure}
 We present an \emph{ablation study} on CoRNN parameters $\epsilon$ and $\gamma$ for Burgers equation in Fig. 7 (\textit{middle}). The contour showcases that the model is robust to the changes in these parameters, highlighted by minor changes in the error throughout the study domain. To showcase the potential of the proposed method for different parameters of the same model, we take the Burgers equation as $u_\mathrm{t} + uu_x = \nu u_{xx}$, with $\nu \in [0.005, 0.05]$ as the parameter. The model is trained for $\nu \in [0.005, 0.036)$ and tested for $\nu \in [0.036, 0.05]$. The solution for $\nu = 0.05$ is presented in Fig. 7 (\textit{right}), showcasing \emph{generalization} capabilities in the parametric space with a relative $L^2$ error of $0.00016$.

\subsection{Allen--Cahn equation}
Fig. 8 presents the contour plot of the approximations of the solution for the Allen--Cahn equation along with snapshots at specific time instances (t = 0.81, 0.99) for RNN and LSTM models.

\begin{figure}[H]
    \centering
    \subfigure[RNN]{\label{fig:AC_pinn_ref_supp2}\includegraphics[width=0.23\textwidth]{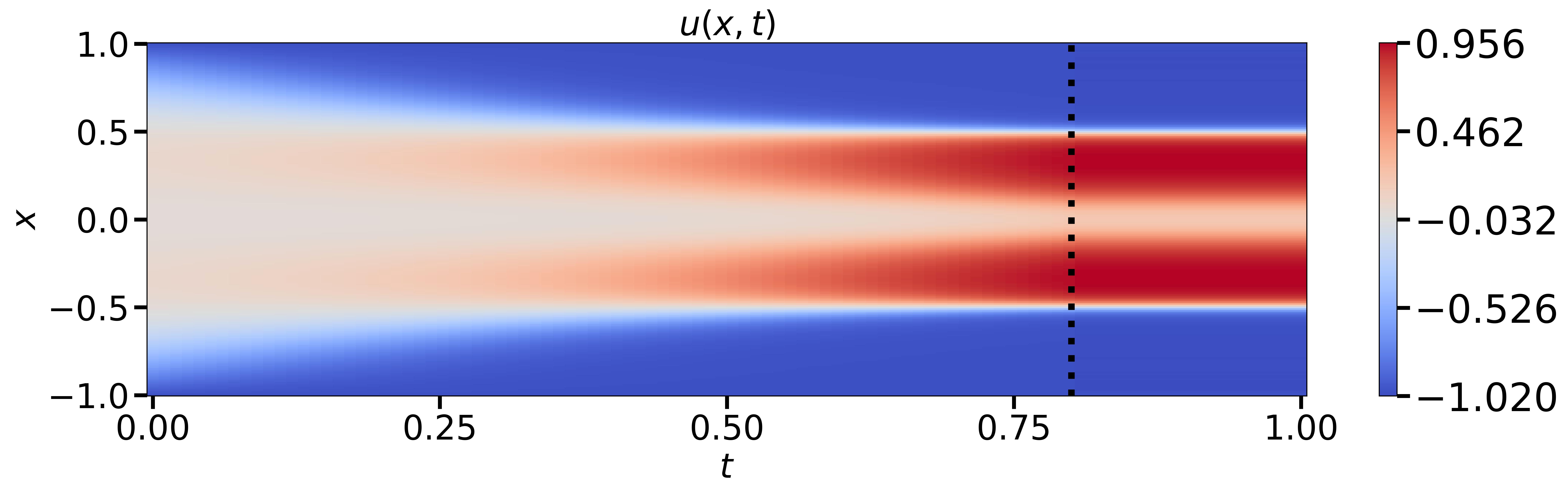}}\hfill
    \subfigure[LSTM]{\label{fig:AC_GRU_contour_supp2}\includegraphics[width=0.23\textwidth]{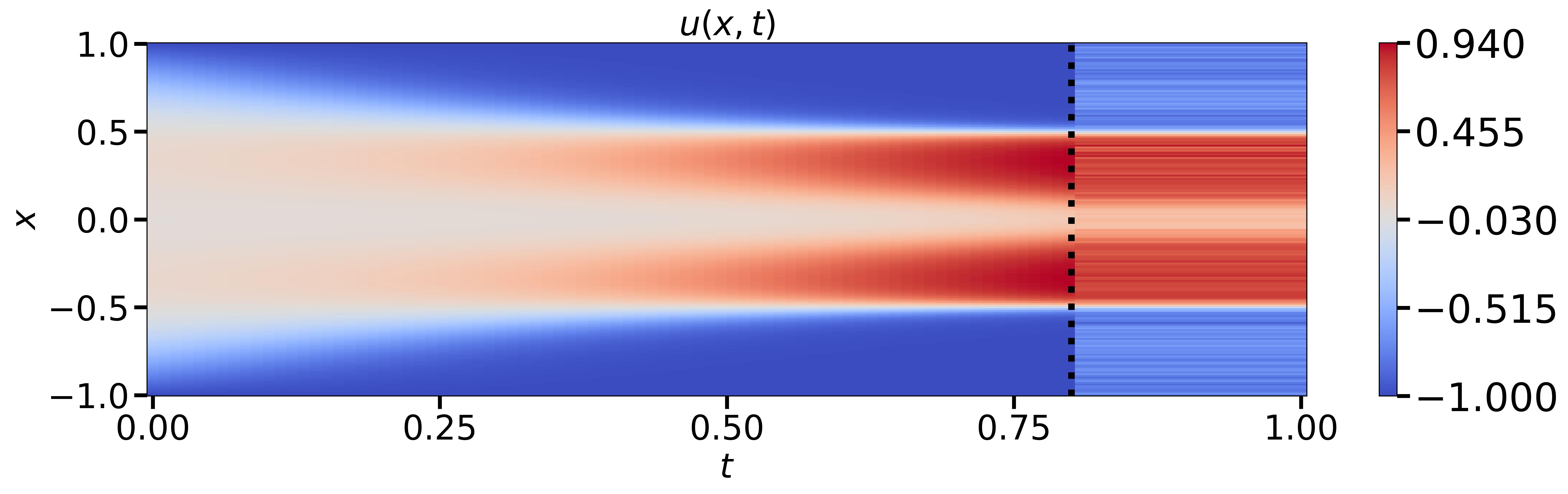}}\hfill
    \subfigure[RNN]{\includegraphics[height=3.5cm, width=3.5cm]{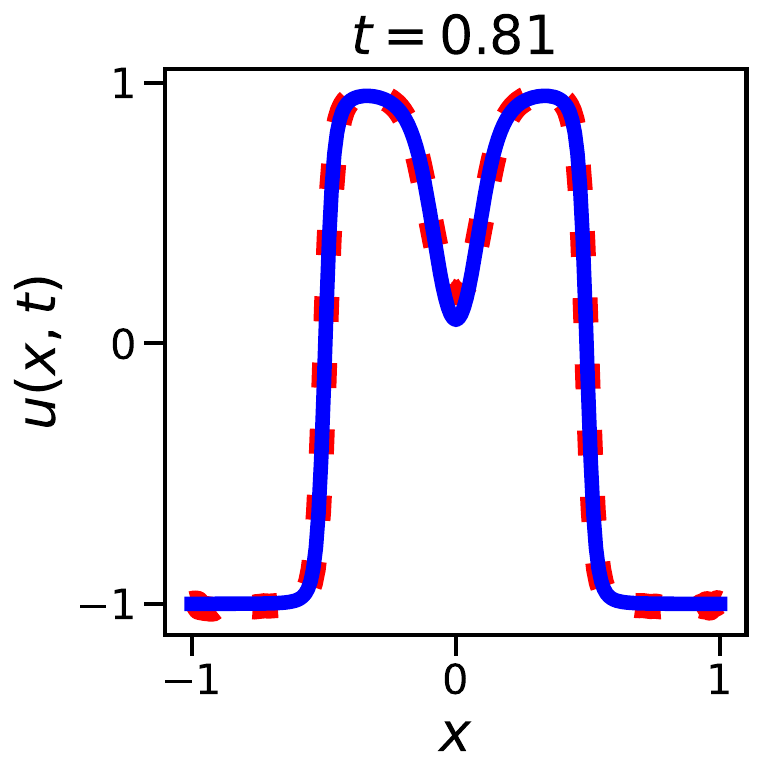}}\hfill
      \subfigure[LSTM]{\includegraphics[height=3.5cm, width=3.5cm]{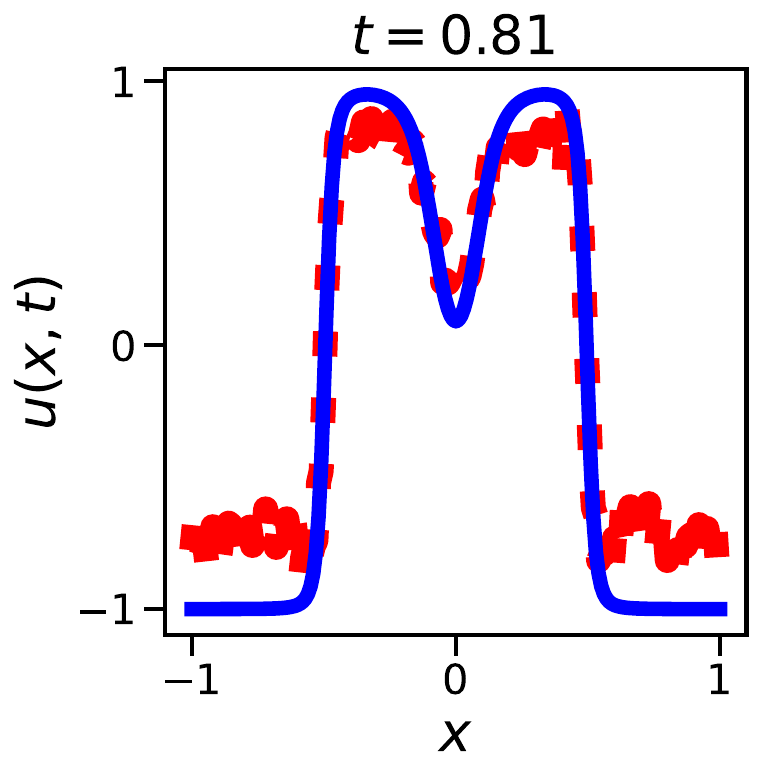}} \hfill
      \subfigure[RNN]{\includegraphics[height=3.5cm, width=3.5cm]{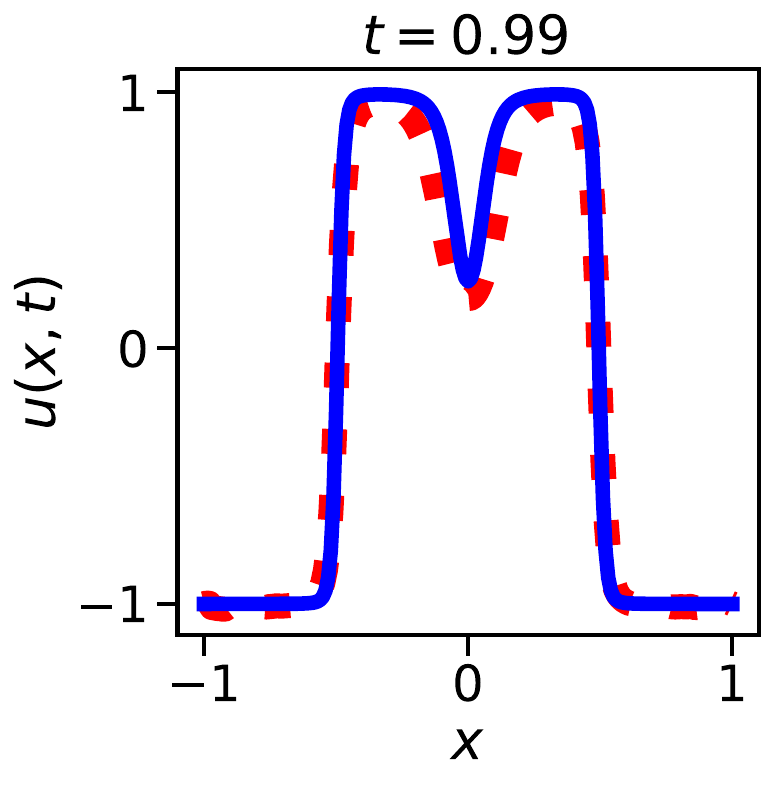}} \hfill
    \subfigure[LSTM]{\includegraphics[height=3.5cm, width=3.5cm]{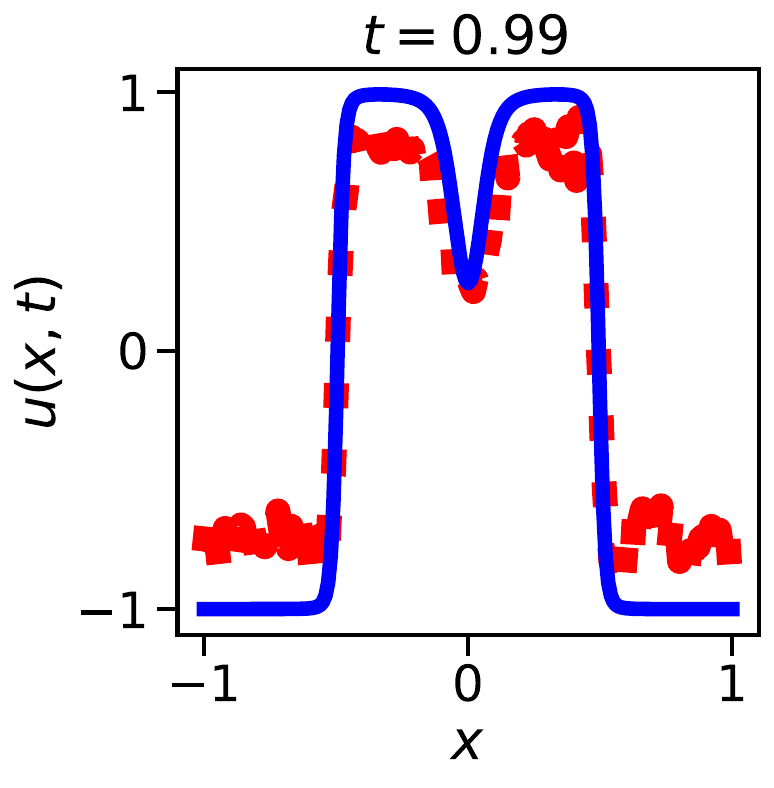}}\hfill
    \caption{Top row: predictions for the Allen--Cahn equation for RNN and LSTM. Bottom row: the solution snapshots at $t=\{0.81, 0.99\}$ obtained in the generalization region.}
\end{figure}

\subsection{Nonlinear Schr\"{o}dinger equation}
Fig. 9 presents the contour plot of the approximations of the solution for the Schr\"{o}dinger equation along with snapshots for specific time instances (t = 1.28, 1.5) for RNN and LSTM models.

\begin{figure}[H]
    \centering
    \subfigure[RNN]{\label{fig:AC_pinn_ref_supp3}\includegraphics[width=0.23\textwidth]{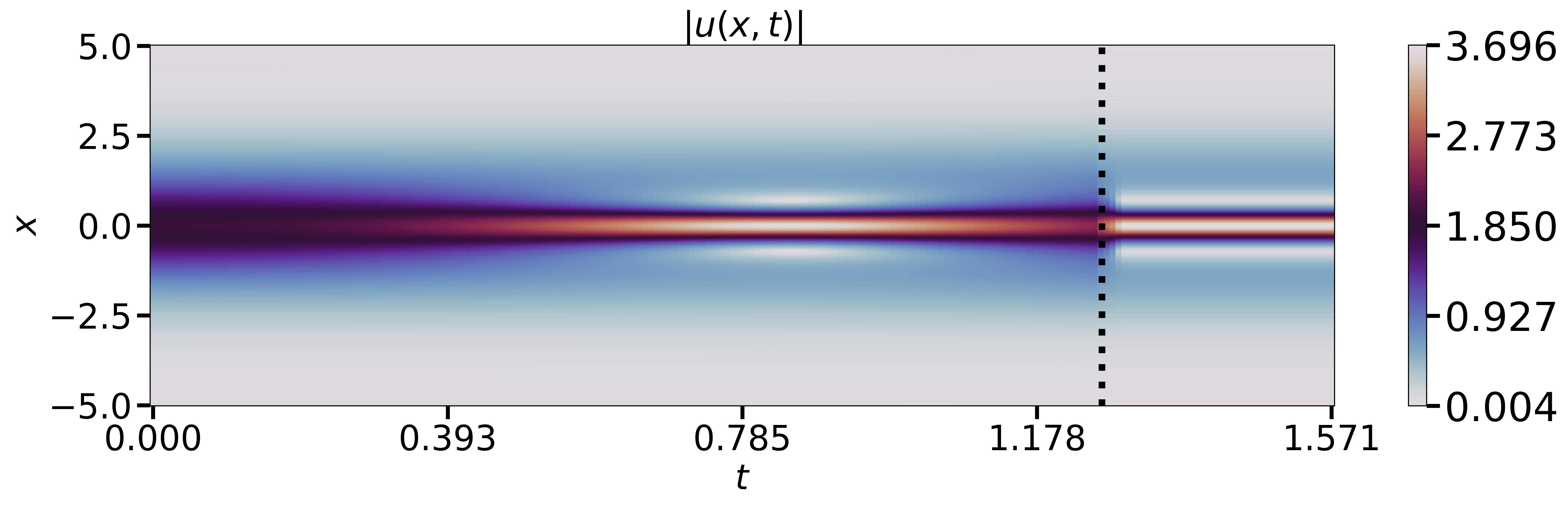}}\hfill
    \subfigure[LSTM]{\label{fig:AC_GRU_contour_supp3}\includegraphics[width=0.23\textwidth]{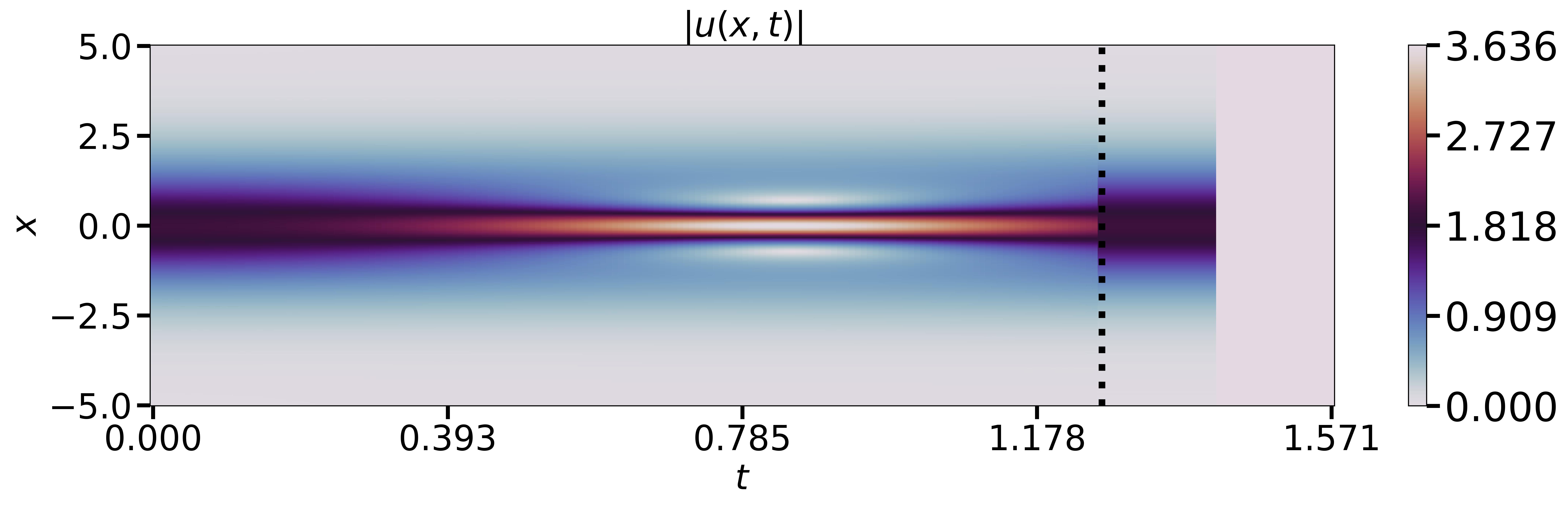}}\hfill
    \subfigure[RNN]{\includegraphics[height=3.5cm, width=3.5cm]{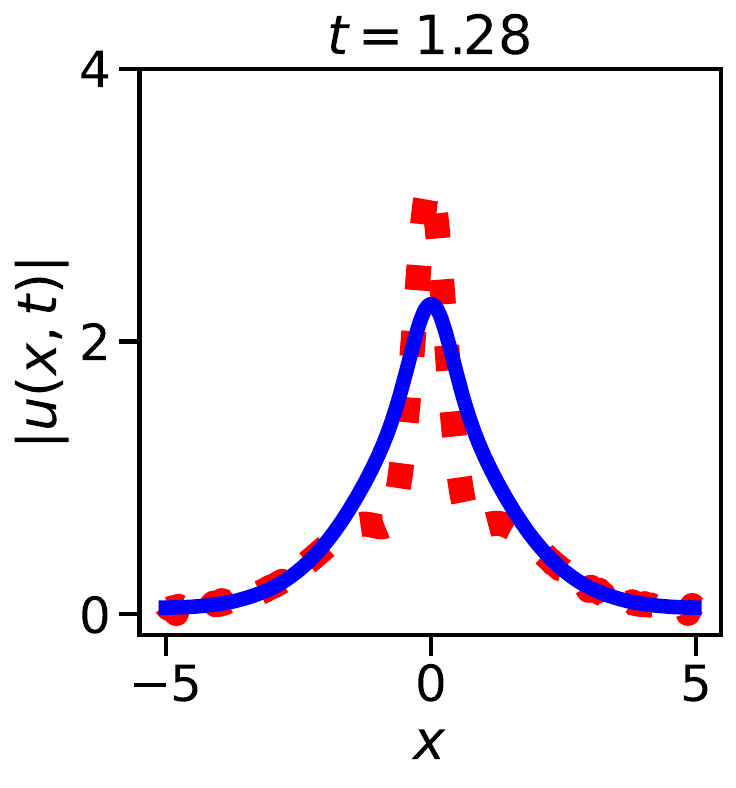}}\hfill
      \subfigure[LSTM]{\includegraphics[height=3.5cm, width=3.5cm]{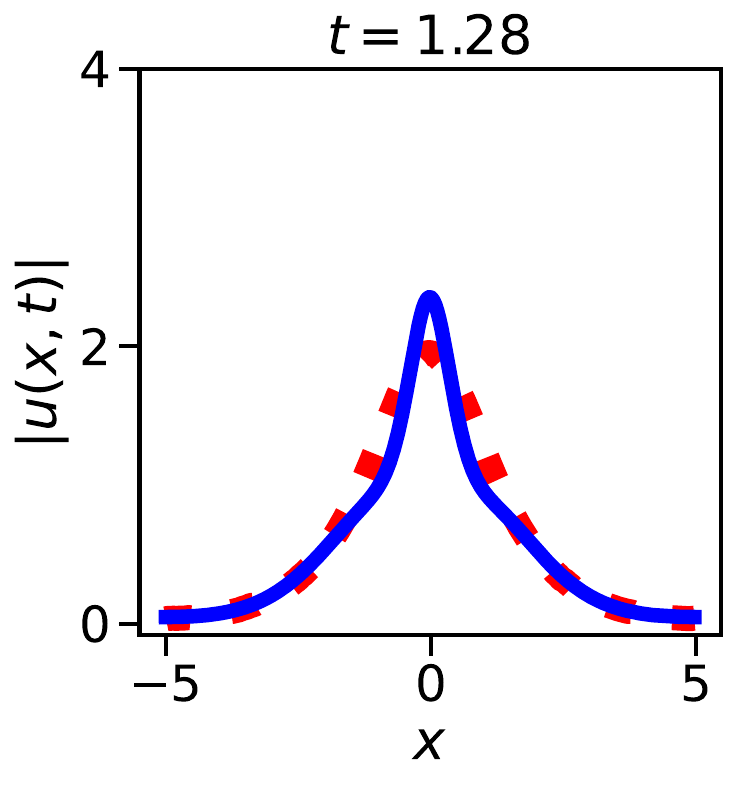}} \hfill
      \subfigure[RNN]{\includegraphics[height=3.5cm, width=3.5cm]{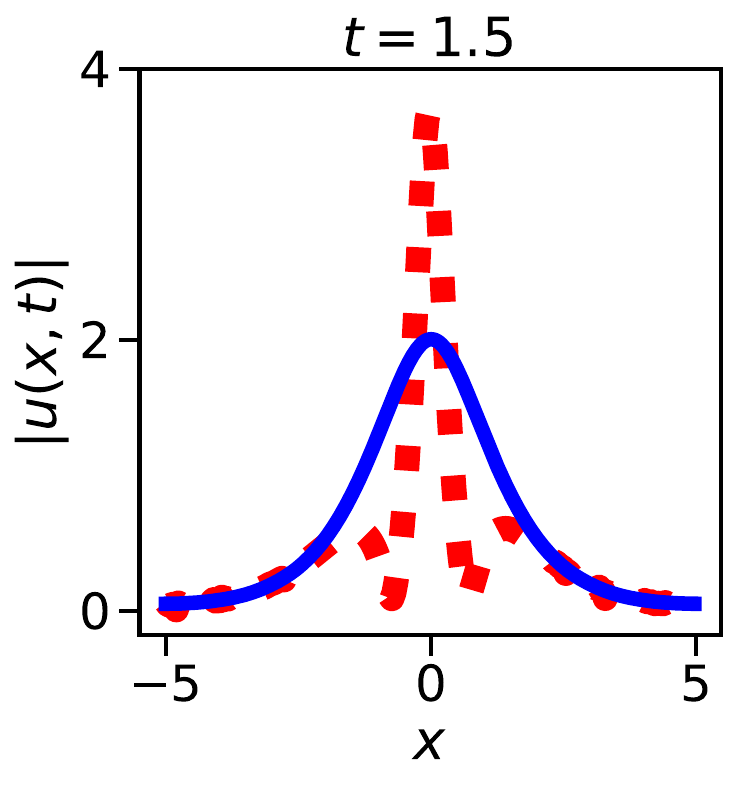}} \hfill
    \subfigure[LSTM]{\includegraphics[height=3.5cm, width=3.5cm]{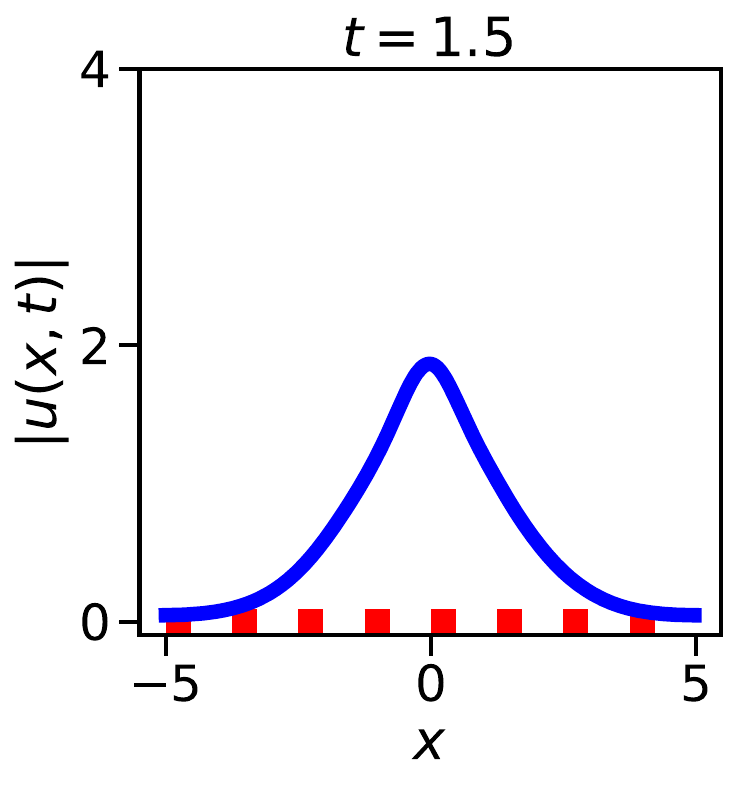}}\hfill
    \caption{Top row: predictions for the Schr\"{o}dinger equation for RNN and LSTM. Bottom row: the solution snapshots at $t=\{1.28, 1.5\}$ obtained in the generalization region.}
\end{figure}

\subsection{Euler--Bernoulli beam equation}
Fig. 10 presents the contour plot of the approximations of the solution for the Euler--Bernoulli beam equation. Also, snapshots for particular time t = 0.83, 0.98 for RNN and LSTM is also presented.

\begin{figure}[H]
    \centering
    \subfigure[RNN]{\label{fig:AC_pinn_ref_supp4}\includegraphics[width=0.23\textwidth]{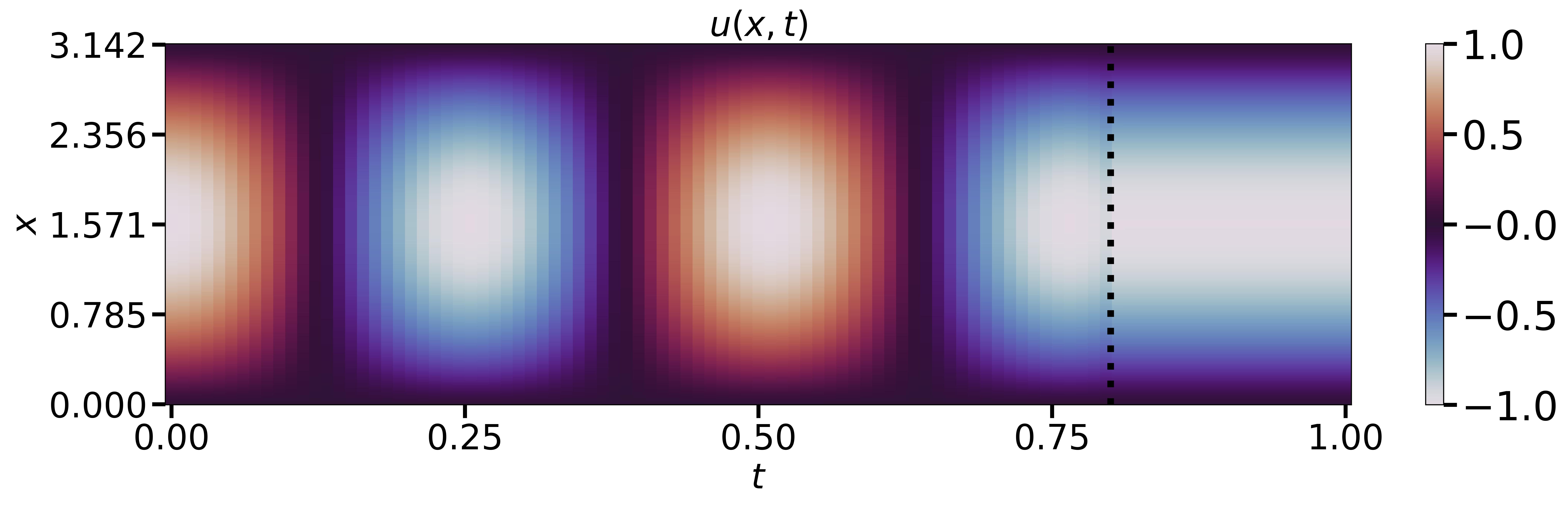}}\hfill
    \subfigure[LSTM]{\label{fig:AC_GRU_contour_supp4}\includegraphics[width=0.23\textwidth]{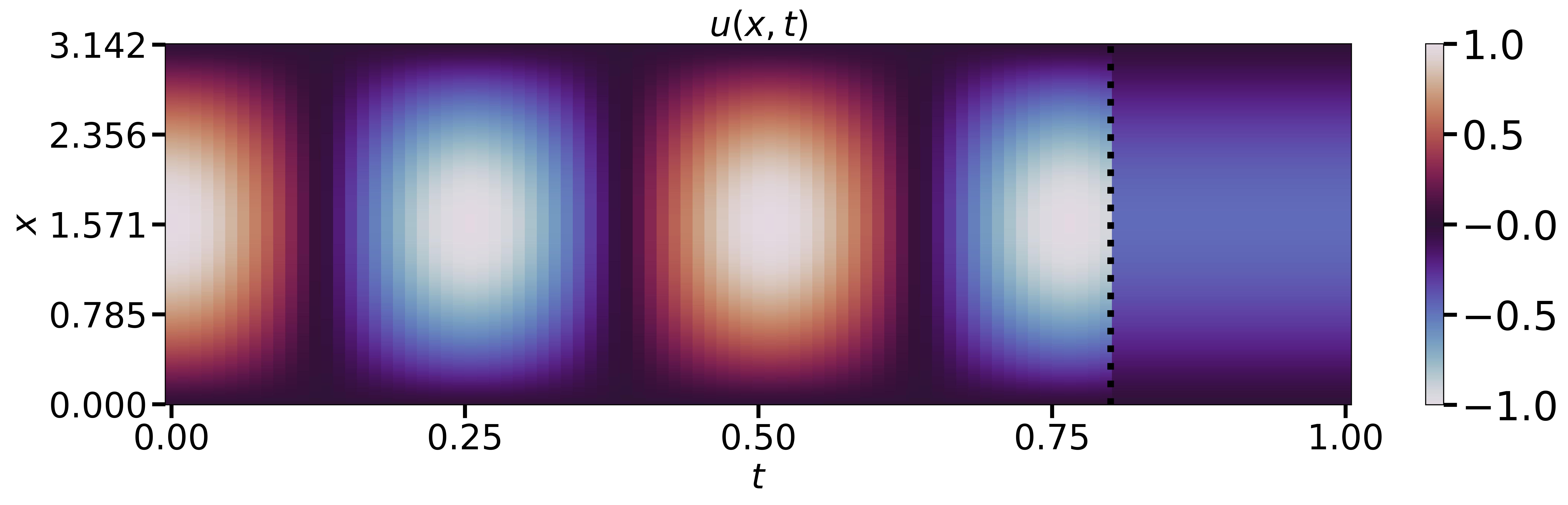}}\hfill
    \subfigure[RNN]{\includegraphics[height=3.5cm, width=3.5cm]{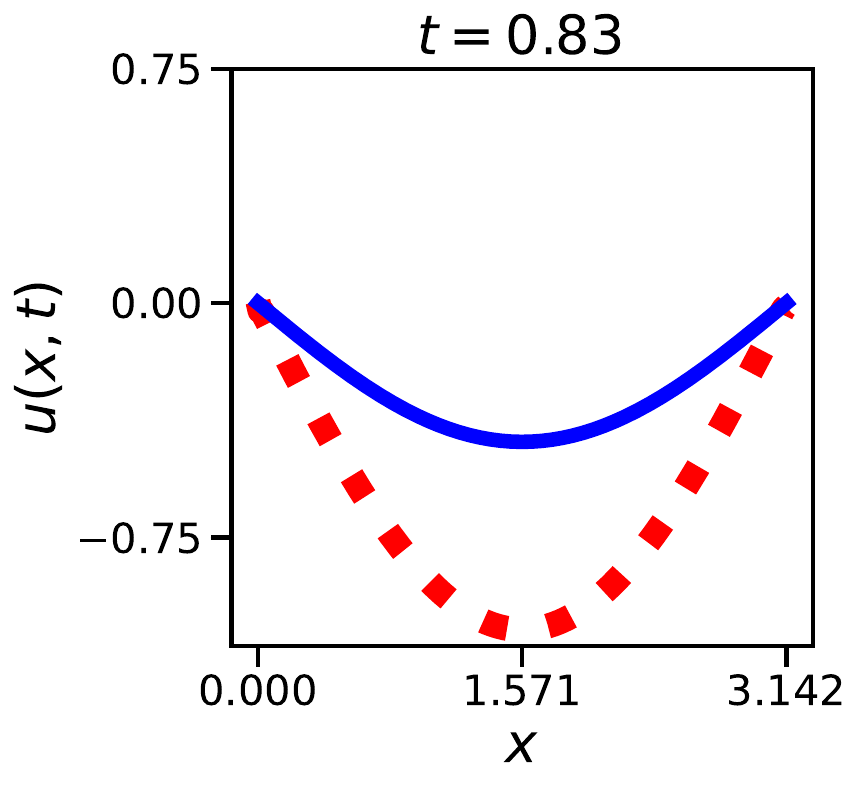}}\hfill
      \subfigure[LSTM]{\includegraphics[height=3.5cm, width=3.5cm]{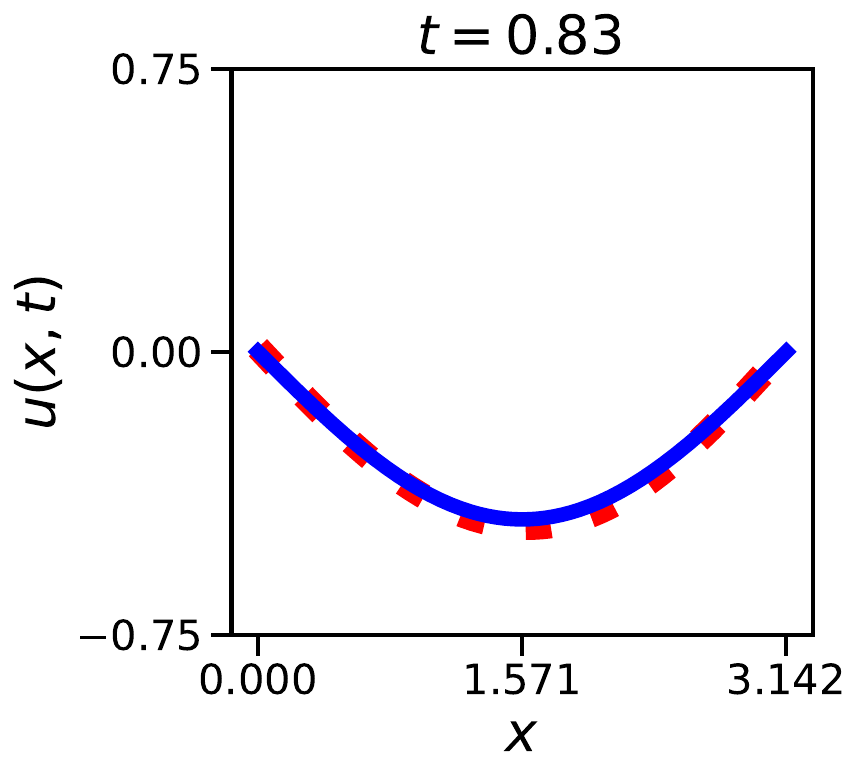}} \hfill
      \subfigure[RNN]{\includegraphics[height=3.5cm, width=3.5cm]{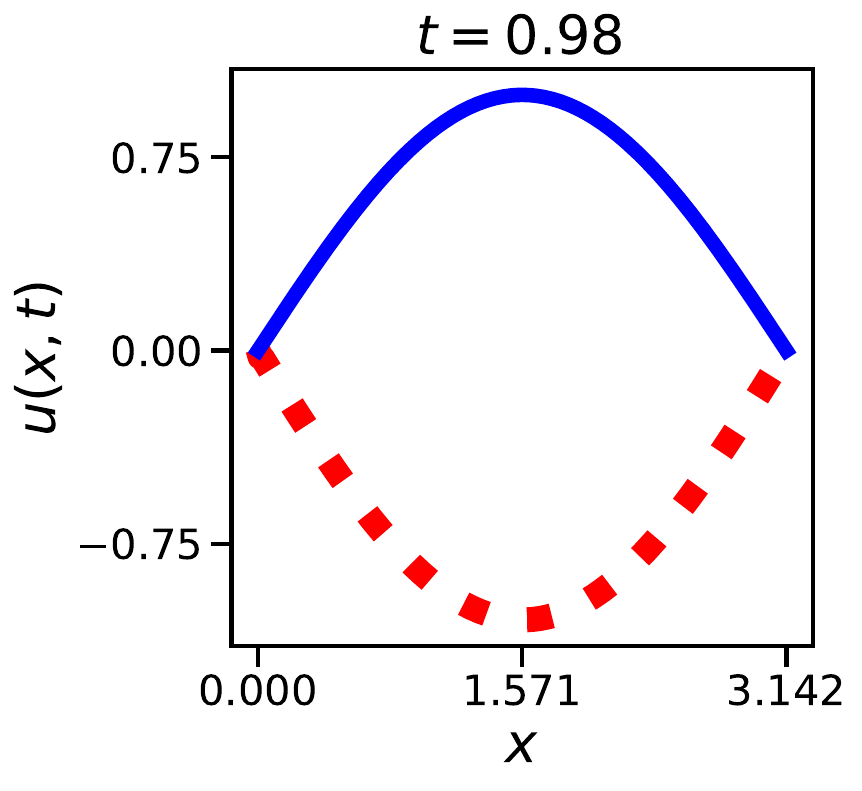}} \hfill
    \subfigure[LSTM]{\includegraphics[height=3.5cm, width=3.5cm]{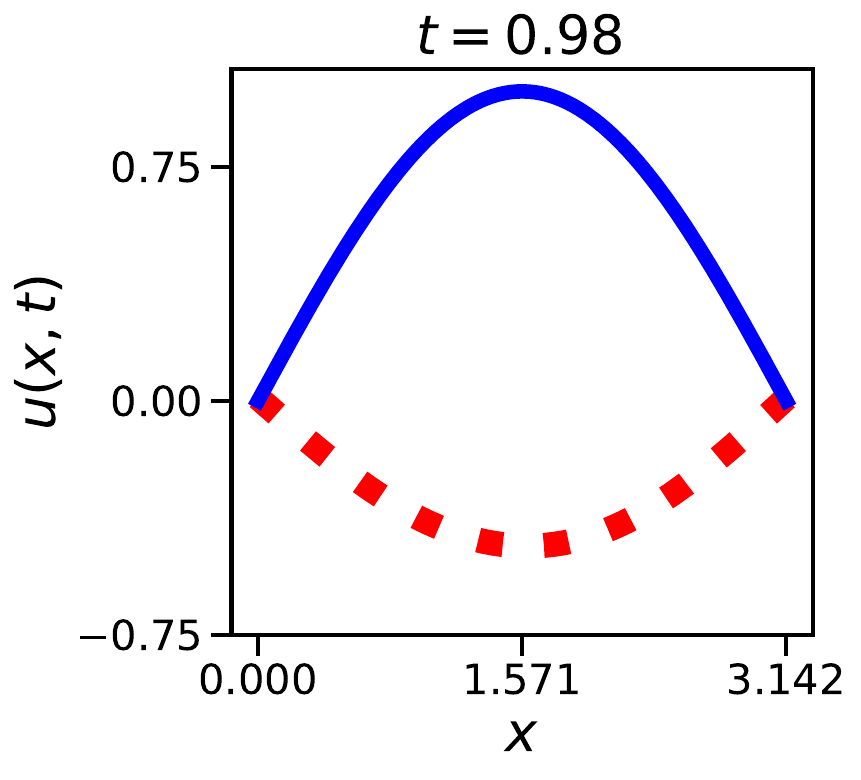}}\hfill
    \caption{Top row: predictions for the Euler--Bernoulli beam equation. Bottom row: the solution snapshots at $t=\{0.83, 0.98\}$ obtained in the generalization region.}
\end{figure}

\subsection{Kovasznay flow}
To showcase the performance of the proposed framework for higher dimensional PDEs, we performed an experiment on 2D Kovasznay flow, taking the problem from \citet{jin2021nsfnets}. Kovasznay flow is modeled by a \emph{higher dimensional system of PDEs} with \emph{complex boundary conditions}. The problem has three unknowns - velocities ($u$, $v$) and pressure ($p$). Fig. 10 shows the reference solution (\textit{left}) and the LEM prediction (\textit{right}) with relative $L^{2}$ error $0.001$ for extrapolating the velocity field.
\begin{figure}[H]
    \centering
    \begin{minipage}[b]{0.48\textwidth}
        \centering
        \includegraphics[width=0.48\textwidth]{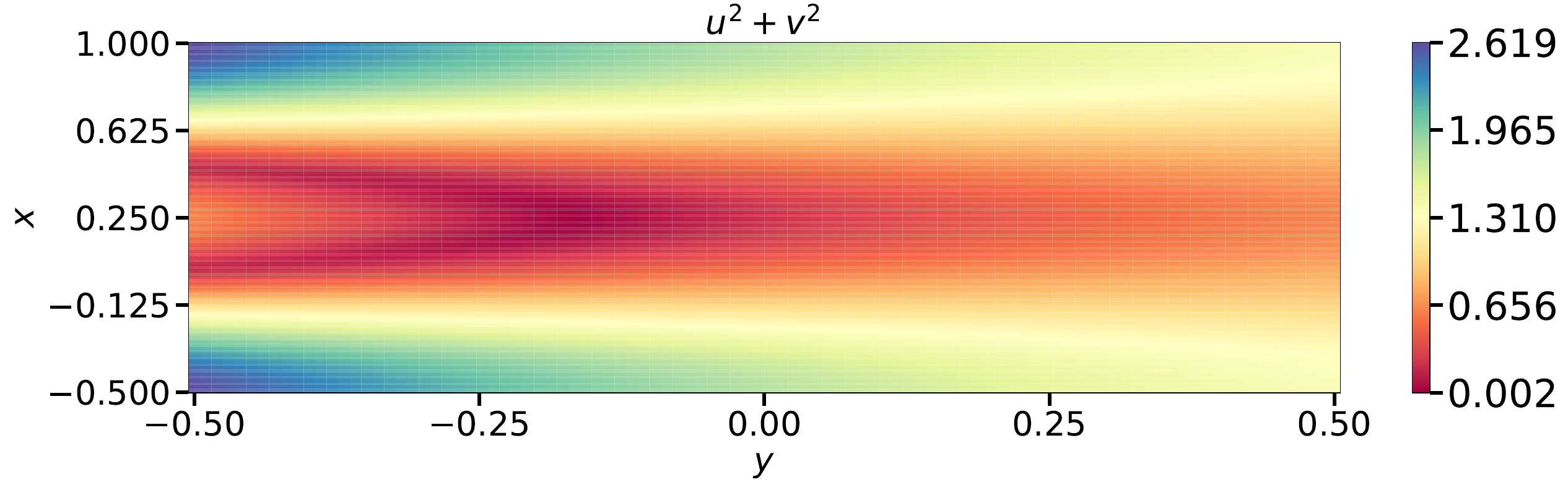}
         \includegraphics[width=0.48\textwidth]{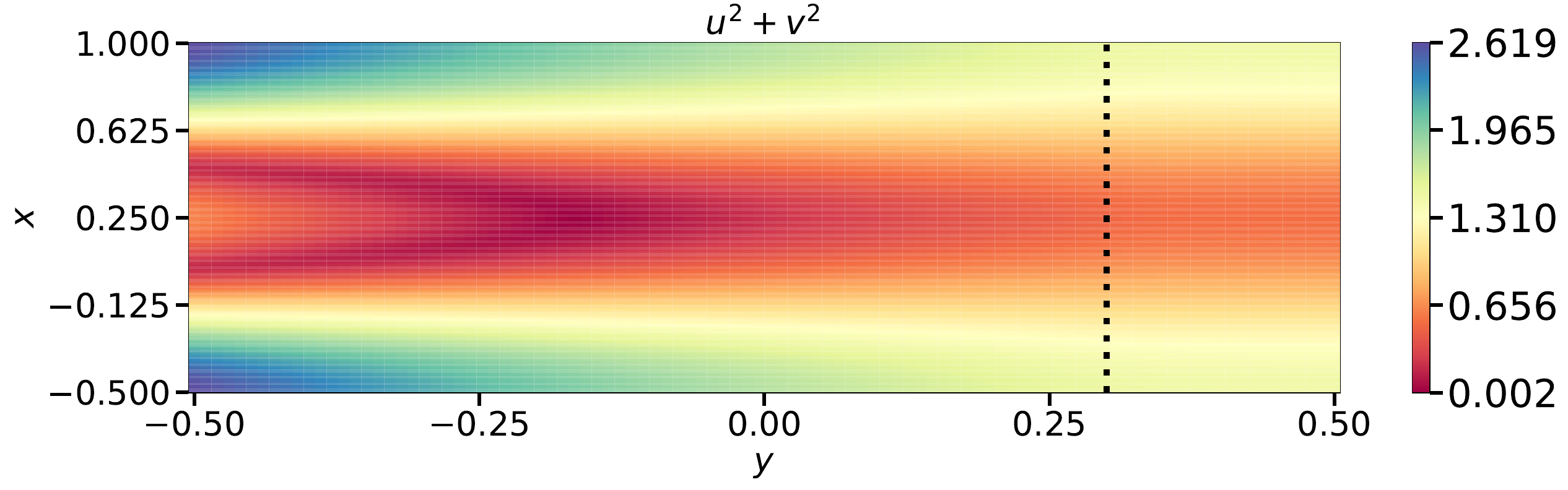}
    \end{minipage}
    \caption{Additional experiment for higher dimensional PDE (Kovasznay flow). Reference (\textit{left}), prediction (\textit{right}).}
    \label{fig:kovaznay}
\end{figure}

\section{D. Error Metrics}
The following subsections present the error metrics utilized within this paper.

\subsection{L2 norm}
The formula for the relative L2 norm in the predicted solution \( \hat{u} \) with respect to the reference solution \( u \) is given by:
\[
\text{Relative L2 norm} = \frac{\|\hat{u} - u\|_2}{\|u\|_2}
\]

where:
\begin{itemize}
  \item \( \|\hat{u} - u\|_2 \) is the Euclidean distance between \( \hat{u} \) and \( u \),
  \item \( \|u\|_2 \) is the Euclidean norm (magnitude) of \( u \).
\end{itemize}

\subsection{Explained variance score}
The formula for the explained variance score is given by:
\[
\text{Explained Variance Score} = 1 - \frac{\sum_{i=1}^{n} (u_i - \hat{u_i})^2}{\sum_{i=1}^{n} (u_i - \bar{u})^2}
\]

where:
\begin{itemize}
  \item \( n \) is the number of testing data points,
  \item \( u_i \) represents the reference solution at the \( i \)-th testing data point,
  \item \( \hat{u_i} \) represents the predicted solution at the \( i \)-th testing data point,
  \item \( \bar{u} \) represents the mean of the reference solution.
\end{itemize}

\subsection{Max error}
The formula for the maximum absolute error is given by:
\[
\text{Max Absolute Error} = \max_{i=1}^{n} |u_i - \hat{u_i}|
\]

where:
\begin{itemize}
  \item \( n \) is the number of testing data points,
  \item \( u_i \) represents the reference solution at the \( i \)-th testing data point,
  \item \( \hat{u_i} \) represents the predicted solution at the \( i \)-th data point,
  \item \( |u_i - \hat{u_i} | \) represents the absolute value of \( u_i - \hat{u_i}  \).
\end{itemize}

\subsection{Mean absolute error}
The formula for the mean absolute error (MAE) is given by:
\[
\text{Mean Absolute Error (MAE)} = \frac{1}{n} \sum_{i=1}^{n} |u_i - \hat{u_i}|
\]

where:
\begin{itemize}
  \item \( n \) is the number of testing data points,
  \item \( u_i \) represents the reference solution at the \( i \)-th testing data point,
  \item \( \hat{u_i} \) represents the predicted solution at the \( i \)-th testing data point,
   \item \( |u_i - \hat{u_i} | \) represents the absolute value of \( u_i - \hat{u_i}  \).
\end{itemize}

\end{document}